\pdfoutput=1

\documentclass[11pt,dvipsnames]{article}
\usepackage{arabtex}
\usepackage{utf8}
\setcode{utf8}
\usepackage[]{ACL2023}

\usepackage{times}
\usepackage{latexsym}

\usepackage[T1]{fontenc}

\usepackage[utf8]{inputenc}
\usepackage{CJKutf8}

\usepackage{microtype}

\usepackage{inconsolata}

\usepackage{graphicx}

\newcommand*{\img}[1]{%
    \raisebox{-.19\baselineskip}{%
        \includegraphics[
        height=0.9\baselineskip,
        width=0.9\baselineskip,
        keepaspectratio,
        ]{#1}%
    }%
}

\usepackage{times}
\usepackage{latexsym}
\usepackage{comment}
\usepackage{caption}
\usepackage[LGR,T1]{fontenc}

\usepackage{hyperref}
\usepackage{url}
\usepackage{multirow}
\usepackage{physics}
\usepackage{array,booktabs}
\usepackage{multirow}
\usepackage{adjustbox}
\usepackage{tabularx,booktabs}
\newcolumntype{Y}{>{\centering\arraybackslash}X}
\newcolumntype{P}{>{\raggedleft\arraybackslash}p{.5cm}}
\usepackage{amsmath}
\usepackage{amsfonts}
\usepackage{enumitem}
\usepackage{xcolor}
\usepackage{colortbl,hhline}
\usepackage[utf8]{inputenc}
\usepackage{soul}

\usepackage{microtype}
\usepackage{graphicx}
\usepackage{placeins}
\usepackage[bottom]{footmisc}

\usepackage{inconsolata}

\usepackage[colorinlistoftodos]{todonotes}

\newcolumntype{L}[1]{>{\raggedright\let\newline\\\arraybackslash\hspace{0pt}}m{#1}}
\newcolumntype{C}[1]{>{\centering\let\newline\\\arraybackslash\hspace{0pt}}m{#1}}
\newcolumntype{R}[1]{>{\raggedleft\let\newline\\\arraybackslash\hspace{0pt}}m{#1}}
\definecolor{green1}{HTML}{D4EFDF}
\definecolor{green2}{HTML}{A9DFBF}
\definecolor{green3}{HTML}{7DCEA0}
\definecolor{green4}{HTML}{52BE80}
\definecolor{red1}{HTML}{FDEDEC}
\definecolor{red2}{HTML}{FADBD8}
\definecolor{red3}{HTML}{F5B7B1}
\definecolor{red4}{HTML}{F1948A}
\usepackage{tikz,graphics,color,fullpage,float,epsf,caption,subcaption}

\title{Frustratingly Easy Label Projection for Cross-lingual Transfer}

\author{Yang Chen, Chao Jiang,  Alan Ritter, Wei Xu \\
  Georgia Institute of Technology \\
 \texttt{\{yang.chen, chao.jiang, alan.ritter, wei.xu\}@cc.gatech.edu} 
\\}

\begin{document}
\maketitle
\begin{abstract}

Translating training data into many languages has emerged as a practical solution for improving cross-lingual transfer. For tasks that involve span-level annotations, such as information extraction or question answering, an additional label projection step is required to map annotated spans onto the translated texts. Recently, a few efforts have utilized a simple mark-then-translate method to jointly perform translation and projection  by inserting special markers around the labeled spans in the original sentence \cite{lewis2020mlqa, hu2020xtreme}. However, as far as we are aware, no empirical analysis has been conducted on how this approach compares to traditional annotation projection based on word alignment. In this paper, we present an extensive empirical study across 57 languages and three  tasks (QA, NER, and Event Extraction) to evaluate the effectiveness and limitations of both methods, filling an important gap in the literature.  Experimental results show that our optimized version of mark-then-translate, which we call EasyProject,  is easily applied to many languages and works surprisingly well, outperforming the more complex word alignment-based methods. We analyze several key factors that affect the end-task performance, and show  EasyProject works well because it can accurately preserve label span boundaries after translation.~\footnote{Our code and data is available at:~\url{https://github.com/edchengg/easyproject}}

\end{abstract}

\section{Introduction}

\begin{figure*}[ht!]
    \centering
    \vspace{-5pt}
\includegraphics[width=0.97\textwidth]{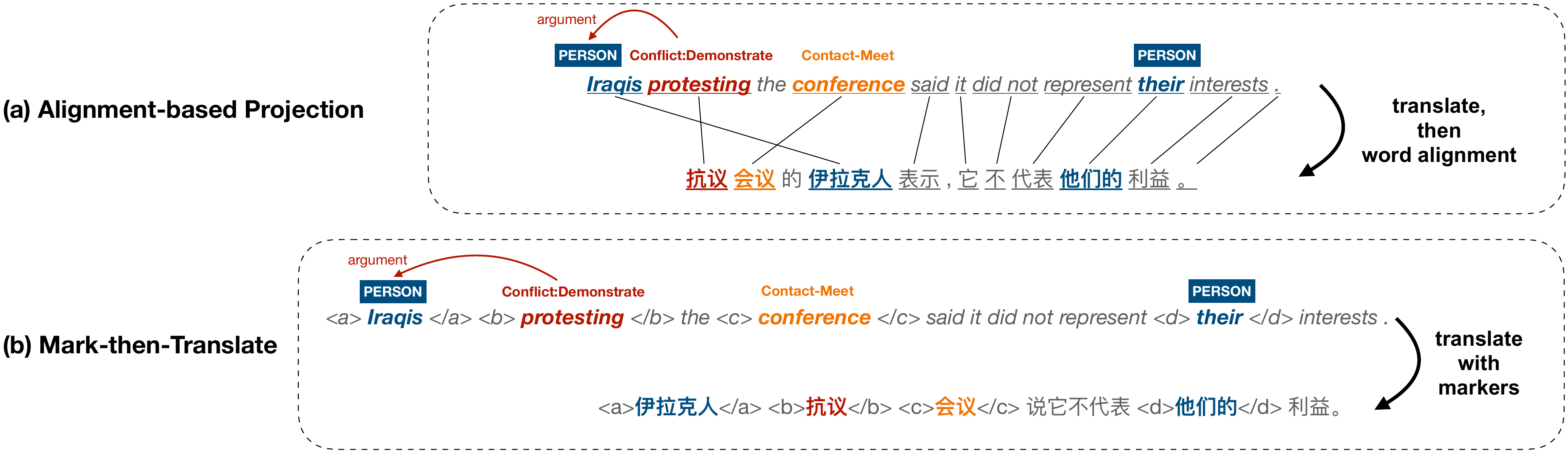}
    \vspace{-6.5pt}
    \caption{Two methods for translating and projecting English ACE event triggers and named entities to Chinese. (a) Pipeline method based on  word alignment: starting with machine translation of the English sentence to Chinese, followed by word-to-word alignment. Then, labeled spans are projected using heuristics. (b) Mark-then-translate: markers are inserted around the  annotated spans in the text. The modified sentence with markers inserted is then fed as input to an MT system, projecting the label span markers to the target sentence as a byproduct of translation.}
\label{fig:EasyProject}
\vspace{-15.5pt}
\end{figure*}

Zero-shot cross-lingual transfer, where models trained on a source language (e.g., English) are directly applied to other target languages, has the potential to extend NLP systems to many languages \cite{nooralahzadeh-etal-2020-zero, keung-etal-2020-dont,chen2021model, niu-etal-2022-onealigner, huang-etal-2022-multilingual-generative}. Yet, its performance still lags behind models that are directly fine-tuned on labeled data (if available) from the target language.
Recent work has shown that combining training data in a source language together with its automatic translation to the target language leads to consistent performance improvements \cite{xue2021mt5,hu2020xtreme}.
However, for NLP tasks that involve span-level annotations, an additional label projection step is needed to map the span annotations onto the translated texts (see Figure \ref{fig:EasyProject}).

Traditionally, this annotation projection step is performed based on word alignment after machine translation \cite{akbik2015generating, aminian-etal-2019-cross}. To avoid the use of complex word alignment models, several recent efforts \cite{lewis2020mlqa, hu2020xtreme} directly translated sentences with span annotations wrapped between special markers (e.g., {\tt <a>} and {\tt </a>}). However, due to limited analysis presented in prior work, it is unclear (1) how well this approach works across different language families,  (2) how robust  MT systems are in handling special markers, as inserting  markers inevitably degrades the translation quality,   and (3) how well marker-based projection works in comparison to traditional alignment-based methods.

In this paper, we present the first systematic study of the mark-then-translate annotation projection technique,  which includes careful evaluation of the choice of markers, projection accuracy, impact on translation quality, robustness to different MT systems, as well as a comparison to traditional alignment-based method across 57 languages (including 18 from Africa)   on 5 datasets and 3 NLP tasks. We also propose an improved variant of marker-based projection, \textsc{EasyProject}, that consistently outperforms the alignment-based approach, while being incredibly easy to use to project a variety of annotations (QA, entities, relations, events) across many languages. The key is to use language-agnostic square bracket markers, combined with an efficient fine-tuning strategy to encourage the multilingual MT system to better preserve the  special markers during  translation.




Our main findings include (1) the marker-based method is surprisingly robust across different translation systems and languages, but the choice of markers matters ($\S$\ref{sec:choice-of-markers}); (2) EasyProject can project annotated spans more accurately and is better at preserving span boundaries than the  alignment-based methods, which is key to its success ($\S$\ref{sec:comparsion_to_alignment}); (3)  fine-tuning an MT system for only 200 steps is sufficient  to improve its robustness in handling special markers during translation ($\S$\ref{sec:easyproject_intro}); (4) the margin of improved cross-lingual transfer is related to the language/script family and amount of pre-training data included in the multilingual model ($\S$\ref{sec:pre-training-data-size}). 
We hope our work will inspire more research on robust models that better handle text markup for the purpose of generating span annotations.


%



\section{Background and Related Work}

\paragraph{Alignment-based Projection.}  Projecting annotations via word alignment typically consists of the following steps: machine translate the available training data into the target language; run word alignment tools on the original and translated sentences; and finally, apply heuristics to map the span-level annotations from the original to translated texts. 
Statistical word alignment tools such as  GIZA++~\citep{giza} and fast-align~\citep{dyer2013fast}  have been widely adopted for projecting part-of-speech tags~\citep{yarowsky-etal-2001-inducing,eskander2020unsupervised}, semantic roles~\citep{akbik-etal-2015-srl,aminian-etal-2017-transferring,daza2020xsrl,fei2020srl}, slot filling~\citep{xu2020end}, semantic parser~\citep{moradshahi-etal-2020-localizing,nicosia-etal-2021-translate-fill}, and NER labels~\citep{ni-etal-2017-weakly,stengel-eskin-etal-2019-discriminative}.
Recent progress on supervised neural aligners~\citep{jalili-sabet-etal-2020-simalign,nagata2020supervised,dou2021awesome,lan2021crfalign} and multilingual contextualized embeddings~\citep{devlin2019mbert,conneau2019xlmr} has further improved alignment accuracy.
However, this pipeline-based method suffers from error propagation, translation shift~\citep{akbik-etal-2015-srl}, and    non-contiguous alignments~\citep{zenkel-etal-2020-end}.
Our analysis in \S \ref{sec:comparsion_to_alignment} shows that the  alignment-based methods are  more error-prone when projecting span-level annotations, compared to the marker-based approaches.

\paragraph{Marker-based Projection.} A few efforts have used mark-then-translate label projection method to translate question answering datasets into other languages ~\citep{lee-etal-2018-semi,lewis2020mlqa, hu2020xtreme,bornea2020qa}. However, the focus of these papers was not the label projection task itself and there was no in-depth analysis on the effectiveness of the approach. For instance, \newcite{lewis2020mlqa} used quotation marks to translate the SQuAD training set into other languages but did not present empirical comparison to any other label projection methods. Similarly, \newcite{hu2020xtreme} used XML tags for the same purpose when creating the XTREME dataset, but this was only briefly mentioned in a few sentences in appendix. Besides QA, MulDA \citep{liu-2021-mulda,zhou2022conner} is a labeled sequence translation method that replaces entities with variable names for cross-lingual NER. However, no comparison with existing  projection methods was presented, as the main focus is generating synthetic labeled data  using language models.

\section{Analysis of Marker-Based Projection}
\label{EasyProject}
The idea of marker-based label projection is straightforward -- wrap labeled spans with special marker tokens, then translate the modified sentence (see an example in Figure \ref{fig:EasyProject}b). The projected spans can be directly decoded from the translation if the markers are retrained. However, inserting markers inevitably degrades the  translation quality.  In this section, we analyze several questions left open by prior work \cite{lewis2020mlqa, hu2020xtreme}, including  (1) how well are the special markers being  preserved in translation, (2) the impact of different marker choices on  translation quality and the performance of cross-lingual transfer.

\subsection{Experimental Setup}
\label{sec:experiment}

We conduct experiments on three NLP tasks and 57 languages with five multilingual datasets to comprehensively evaluate  the marker-based method. Most multilingual datasets are created  by either (1) directly collecting and annotating data in the target language, or (2) translating English data with human or machine translation and then projecting labels manually or automatically.  Four of our selected datasets  were created with the first method, as evaluation on the translated datasets may over-estimate performance on a target language, when in fact a model might only perform well on  {\em translationese} \citep{riley2020translationese}. 





%
\paragraph{Datasets.} Our experiments include NER via the WikiANN~\citep{pan2017wikiann,rahimi2019ner} and MasahkaNER 2.0 \cite{adelani2022masakhaner} datasets (\S \ref{sec:comparsion_to_alignment}), in addition to CoNLL-2002/2003 multilingual NER datasets \cite{sang2002conll,sang2003conll} for comparison with \citet{liu-2021-mulda} (\S \ref{sec:mulda}).  
For event extraction, we use the ACE05 corpus \cite{walker2006ace},  which consists of six sub-tasks: entity and relation extraction, event trigger/argument identification and classification.
For QA, we use TyDiQA-GoldP \cite{clark-etal-2020-tydi}, which contains challenging questions written in eight languages.
Data statistics are shown in Table \ref{table:stat} and \ref{table:stat_conll} in Appendix.

\renewcommand{\arraystretch}{1.25}
\begin{table}[t!]
\centering
\small
\resizebox{\linewidth}{!}{
\begin{tabular}{@{\hspace{0.05cm}}l@{\hspace{0.01cm}}c@{\hspace{0.05cm}}c@{\hspace{0.06cm}}c@{\hspace{0.08cm}}c@{\hspace{0.1cm}}}
\toprule
& \textbf{WikiANN} & \textbf{MasakhaNER} & \textbf{ACE05} & \textbf{TyDiQA} \\
\midrule
\# of Lang. & 39 & 20 (from Africa) & 2  & 8  \\ 
\# of Docs & -- & -- & 526/31/40 & 3,696/440/-- \\ 
\# of Sent. & 20k/10k/10k & 4.4k/638/1.2k  & 19k/901/676 & 17k/2,122/-- \\
Avg.  Length & --/8.0 & --/23.9   & 519.3/14.9 & 96.8/21.0 \\
Avg.  \# of Spans & 1.4 & 1.8 & 2.9 & 1.0 \\
\bottomrule
\end{tabular}
}
\vspace{-5pt}
\caption{The detailed statistics of train/dev/test sets for each dataset.  \textbf{Avg. Length} represents the average number of tokens in each article/sentence, and \textbf{Avg.  \# of Spans} denotes the average number of annotated spans in each sentence (in each article for TyDiQA).} 
\vspace{-15pt}
\label{table:stat}
\end{table}



\paragraph{IE and QA Models.} We use  XLM-RoBERTa$_\text{large}$ ~\citep{conneau2019xlmr} as the backbone  model, except where noted.\footnote{We also experiment with mT5$_{large}$, mT5$_{\text{XL}}$, and mT5$_{\text{XXL}}$~\citep{xue2021mt5} on WikiANN-NER (Table \ref{table:mt5}). }
For NER and QA, we fine-tune XLM-R with standard tagging and SQuAD-style span prediction layers. For event extraction, we use the OneIE framework~\citep{lin2020oneie}, a joint neural model for information extraction with global features. We report average F$_1$ scores over three runs with different random seeds. More implementation details can be found in Appendix \ref{appendix:implementation_details_IE}.

\paragraph{MT Systems.} We experiment with two MT systems: (1) the Google Translation (GMT) API,\footnote{\url{https://cloud.google.com/translate}} and (2) an open-sourced multilingual  translation model NLLB (No Language Left Behind)~\citep{nllb2022} with 3.3 billion parameters, supporting the translation between any pair of 200 languages.\footnote{\url{https://github.com/facebookresearch/fairseq/tree/nllb}} 


\subsection{Choice of Markers}
\label{sec:choice-of-markers}
Ideally, a good span marker should minimize the impact on translation quality while having a high chance of being preserved during translation. However, prior works used quotation marks (`` '')~\citep{lee-etal-2018-semi,lewis2020mlqa} and XML/HTML tags (e.g., <a> or <PER>)~\citep{hu2020xtreme, ahmad2021gate} without much justification, which we address below.


\paragraph{Preserved in Translation.}  In a pilot study, we experimented with several markers, including XML tags, \texttt{[]}, \text{`` ''}, \texttt{()}, \texttt{<>}, and \text{\{\}}, etc. We found that both MT systems work reasonably well to retain square brackets (\texttt{[]}) and XML markers during the translation across many languages, while other markers that have language-specific formats are easily lost in translation.  For example, quotation marks (`` '') are often translated in a language-specific way, e.g., «» in Russian, and  are sometimes  lost entirely in Arabic and Finnish, leading to low projection rates: 53\% for Russian, 76\% for Arabic, and 79\% for Finnish based on TyDiQA dataset. The \textit{projection rate} is measured by the percentage of data in which the numbers and type of special  markers in the translations match with the source sentences. To improve the robustness of MT system in handling  markers, we found  further fine-tuning the MT system on synthetic data, where the special markers are inserted around  name entities in parallel sentences,  for only 200 steps is sufficient to boost the projection rate while maintaining translation quality (more details in $\S$\ref{sec:easyproject_intro}).

%

\renewcommand{\arraystretch}{1.1}
\begin{table}[b!]
\centering
\small
\vspace{-11pt}
\begin{tabularx}{\linewidth}{llrPPP}
\toprule
\multirow{2}{*}{\textbf{\textit{en} $\rightarrow$ Lang.}}  & \multirow{2}{*}{\textbf{Corpus}} & \multirow{2}{*}{\# \textbf{sent}} & \multicolumn{3}{c}{GMT - \textbf{BLEU}}\\
\cmidrule{4-6}
&& & Orig. & XML & \texttt{[]}\\
\midrule
Arabic (\textit{ar}) & TED18 & 1,997 & \textbf{20.7} &14.0& \underline{15.1}\\
German (\textit{de}) & TED18 & 1,997 & \textbf{44.5} &33.9 & \underline{41.9}\\
Spanish (\textit{es}) & TED18&1,997 & \textbf{45.9} &34.2 & \underline{35.4} \\
French (\textit{fr}) & TED18&1,997 & \textbf{37.6} &31.0  &\underline{31.9}\\
Hindi (\textit{hi}) & TED18&1,070 & \textbf{14.5} &12.8 & \underline{13.0}\\
Russian (\textit{ru}) & WMT&1,997 & \textbf{36.4} &28.5 & \underline{35.2}\\
Vietnamese (\textit{vi}) & TED18&1,997 & \textbf{32.8} & \underline{28.5} &27.0\\
Chinese (\textit{zh}) & WMT&1,997 & \textbf{40.6}	&33.4 & \underline{37.1}\\
\midrule
AVG &  & 1,881 & \textbf{34.1} &27.0 & \underline{29.6}\\
\bottomrule
\end{tabularx}
\vspace{-5pt}
\caption{Comparsion of translation  quality with different span markers, where the \textbf{best} and \underline{second best} are marked. Overall, square brackets (\texttt{[]}) have less negative impact   compared to XML tags. ``Orig.'' denotes the translation when no marker is inserted.}
\vspace{-5pt}
\label{table:bleu_corpus}
\end{table}
%

\paragraph{Impact on Translation Quality.} After narrowing down the choices to XML tags and square brackets, we further measure the impact of adding markers on the translation quality by adopting the evaluation setup used by \citet{fan2021beyond}. We compare translation quality, with and without markers inserted, from English to various target languages using  BLEU score. Table~\ref{table:bleu_corpus} presents the experimental results with Google Translation. Examples of errors are shown in Table \ref{tab:translation_error}.  We find that inserting special  markers indeed  degrades translation quality, but overall,  square brackets have less negative impact compared to XML tags.    We hypothesize this is because using \texttt{[]} introduces less number of extra subword tokens  in the encoding and decoding of the text during translation, compared to XML tags. More results on 55 languages using the NLLB  translation system and more details about the evaluation setup can be found in Appendix \ref{appendix:translation-quality-evaluation}. 




\begin{figure*}[t!]
    \centering
    \vspace{-5pt}
    \includegraphics[width=\textwidth]{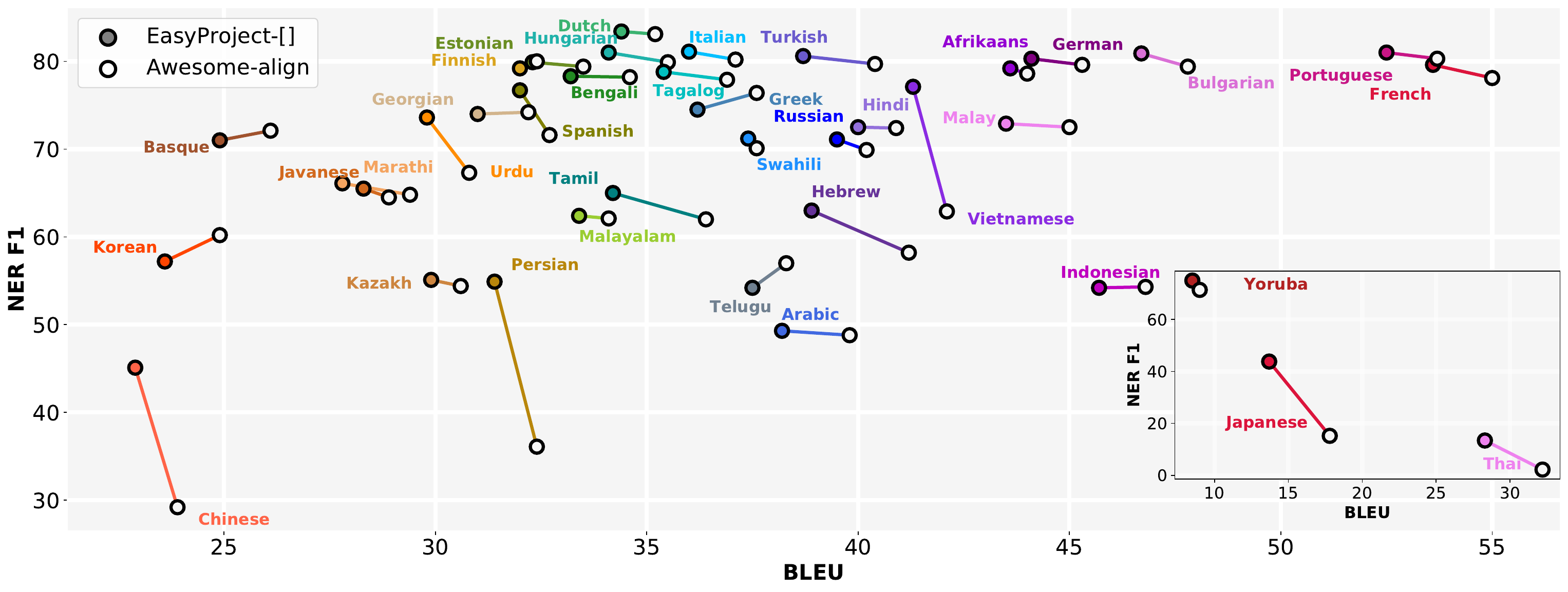}
    \vspace{-25pt}
    \caption{Comparison of translation quality (x-axis) and end-task performance (y-axis) for different label projection methods on the WikiANN dataset using NLLB translation system. EasyProject\img{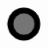}(\S\ref{sec:easyproject_intro}) outperforms alignment-based approach\img{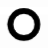}on F$_1$ scores for most languages, though inserting  markers degrades translation quality. The experimental setting is detailed in \S \ref{sec:comparsion_to_alignment}.}
    \label{fig:bleu_f1}
    \vspace{-15pt}
\end{figure*}

\begin{CJK*}{UTF8}{gbsn} 
\renewcommand{\arraystretch}{1.1}
\begin{table}[t!]
    \small
    \centering
    \resizebox{\linewidth}{!}{
    \begin{tabular}{@{\hspace{0.025cm}}r@{\hspace{0.025cm}}l@{\hspace{0.025cm}}}
    \toprule

       
       
       \textbf{English} \textbf{\#1:}   &  The \underline{divorce} settlement called for \underline{Giuliani} to \underline{\smash{pay}} \underline{Hanover}\\
       &more than \$6.8 million, according to the \underline{\smash{reporter}} . \\
       \midrule
       \textbf{Orig.:}   & 据\underline{记者}称，\underline{离婚}协议要求\underline{朱利安尼}向\underline{汉诺威} \\
       & \underline{支付}超过680万美元。 \\
       \midrule
       \textbf{\texttt{[}\,\texttt{]}\,:}  & 据\textcolor{blue}{[}记者\textcolor{blue}{]}{\setlength{\fboxsep}{1pt}\colorbox{Blue!25}{报道}}，\textcolor{blue}{[}离婚\textcolor{blue}{]}协议要求\textcolor{blue}{[}朱利安尼\textcolor{blue}{]}\textcolor{blue}{[}支付\textcolor{blue}{]}  \\
       & \textcolor{blue}{[}汉诺威\textcolor{blue}{]}超过680万美元。 \\
        \midrule
       \textbf{XML:}   & 据\textcolor{red}{\text{<e>}}{\setlength{\fboxsep}{1pt}\colorbox{Red!25}{记者}}\textcolor{red}{\text{<}}，\textcolor{blue}{\text{<a>}}离婚\textcolor{blue}{\text{</a>}}和解协议要求\textcolor{blue}{\text{<b>}}朱利安尼 \\
       & \textcolor{blue}{\text{</b><c>}}支付\textcolor{blue}{\text{</c><d>}}汉诺威\textcolor{blue}{\text{</d>}}超过680万美元。\textcolor{red}{\text{/e>} 。} \\
        \midrule
        \midrule
        \textbf{English} \textbf{\#2:}  &  The \underline{WTO}  is headquartered in \underline{Geneva} .  \\
        \midrule
       \textbf{Orig.:}  & \multicolumn{1}{r}{.
       \underline{\smash{\<جنيف>}} \<في> \underline{\smash{\<لمنظمة التجارة العالمية >}}  \<يقع المقر الرئيسي>
       } \\
       \midrule
       \textbf{\texttt{[}\,\texttt{]}\,:}  &  \multicolumn{1}{r}{. \textcolor{blue}{[} \<جنيف> \textcolor{blue}{]} \textcolor{red}{\underline{\smash{\<لمنظمة التجارة العالمية>}}} \<يقع المقر الرئيسي  في> } \\

        \midrule
    \textbf{XML:}  & \multicolumn{1}{r}{. \textcolor{blue}{</b>}  \< جنيف> } \textcolor{blue}{<b>}  \<في> \textcolor{blue}{</a>}  WTO \textcolor{blue}{<a>} \<يقع المقر الرئيسي ل>  \\
       \bottomrule
       
    \end{tabular}}
    \vspace{-7.5pt}
    \caption{ Example  \textcolor{red}{errors}  and \textcolor{blue}{correctly projected} markers  with GMT. In \#1, a necessary Chinese verb ``{\setlength{\fboxsep}{1pt}\colorbox{Blue!25}{报道}} (report)'' is lost in the XML-marked translation, while tags (\textcolor{red}{<e>, /e>}) are also mismatched due to the word reordering of ``{\setlength{\fboxsep}{1pt}\colorbox{Red!25}{记者}} (reporter)''. In \#2, \texttt{[]}-marked translation fails to preserve the square brackets (\texttt{[]}) around the Arabic translation of ``WTO'' (marked by \textcolor{red}{\underline{underline}}).}
    \label{tab:translation_error}
    \vspace{-7.5pt}
\end{table}
 \end{CJK*}

\renewcommand{\arraystretch}{1.1}
\begin{table}[t!]
\centering
\small
\begin{tabular}{lcccc}
\toprule

\multirow{2}{*}{\textbf{\textit{en} $\rightarrow$ Lang.} }  & \multirow{2}{*}{\citeauthor{hu2020xtreme}} & \multicolumn{3}{c}{GMT - \textbf{TyDiQA F$_1$}}\\
\cmidrule{3-5}
&   & XML & \texttt{[]} & `` ''\\

\midrule
Arabic (\textit{ar}) & \underline{68.8}	& 68.4&	\textbf{71.7}&66.5\\
Bengali (\textit{bn}) &58.6	&\underline{64.8}	&64.1 & \textbf{69.3}\\
Finnish (\textit{fi}) &69.4	&\underline{69.6}	&\textbf{70.8}&68.0\\
Indonesian (\textit{id}) &75.5	&76.0	&\textbf{78.6}&\underline{77.3}\\
Korean (\textit{ko}) & 56.8	&55.6	&\underline{59.0}&\textbf{59.6}\\
Russian (\textit{ru}) &49.5	& \underline{65.7}	&\textbf{66.1}&52.3\\
Swahili (\textit{sw}) &69.1	&\textbf{70.4}	&\underline{70.1} & \underline{70.1} \\
Telugu (\textit{te}) & \textbf{70.2}	& \underline{69.0}	&67.3 & 67.9\\
\midrule
AVG & 64.7	&\underline{67.4}	&\textbf{68.5}& 66.4\\
\bottomrule
\end{tabular}
\vspace{-4pt}
\caption{Comparison of different markers on TyDiQA-GoldP by training on the translated \textit{projected data only}. Overall, square brackets (\texttt{[]}) have the best transfer learning performance.}
\vspace{-17pt}
\label{table:quotes}
\end{table}

\paragraph{Impact on Transfer Learning.} We next evaluate the impact of different marker choices on the performance of cross-lingual transfer.  The results on the TyDiQA dataset are presented in Table~\ref{table:quotes}.  On average, square brackets (\texttt{[]}) have the best transfer learning performance. We also  directly compare with the projection data released by  \citeauthor{hu2020xtreme}, which utilizes XML tags and a  Google internal translation system in the year 2020 to translate  QA datasets. More results on NER and event extraction tasks, and comparison with the alignment-based  projection methods are presented in Table \ref{table:main_results}.



\vspace{-2pt}
\section{\textsc{EasyProject}} 
\label{sec:easyproject_intro}
\vspace{-1pt}


 Based on our analysis, we develop an optimized version of the mark-then-translate method, which we call \textsc{EasyProject}.\footnote{This name was inspired by \citet{daume2007frustratingly}.}   Our improvements target the two weaknesses of the marker-based approach: (1) special markers may get lost during translation; and (2) although  square brackets (\texttt{[]})  show  strong performance, they don't carry the correspondence between original spans and the ones in the translation (e.g., \texttt{[} Churchill \texttt{]} was born in \texttt{[} England \texttt{]} in \texttt{[} 1874 \texttt{]}.), as the XML tags (e.g., \texttt{<a>} Churchill \texttt{</a>} was born in \texttt{<b>} England \texttt{</b>} in \texttt{<c>} 1874 \texttt{</c>}.). If multiple annotated spans with different labels exist in one sentence, it is challenging to assign labels to the  projected entities in the translation, as word order can change between languages. 

%



\subsection{Fine-tuning NLLB}
To  improve the robustness of the MT system in handling special markers, we further fine-tuned the NLLB  model on synthetic data. We utilize parameter-efficient fine-tuning by only updating the last layer of the encoder and decoder, which take 4.2\% of all parameters. We found fine-tuning 200 steps is sufficient to improve the projection rate on  TyDiQA dataset from  70\% to 96.4\% while maintaining the  translation quality.

\label{sec:fine-tune-nllb}

\begin{figure}[t!]
    \centering
    \vspace{-3pt}
    \includegraphics[width=\linewidth]{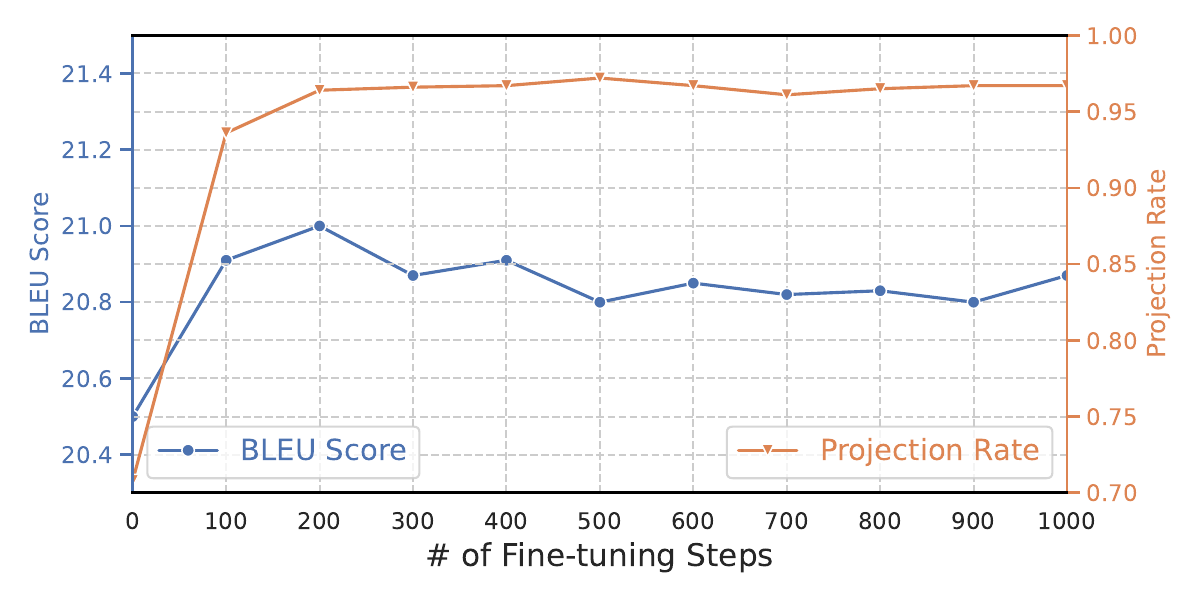}
    \vspace{-25pt}
    \caption{The changes of   projection rate and translation quality (measured by BLEU score)  as fine-tuning more steps. We found  200 steps are sufficient to boost the projection rate while maintaining  translation quality.}
    \label{fig:projection-rate-translation-quality-change}
    \vspace{-15pt}
\end{figure}

\paragraph{Creating Synthetic Data.}
We first construct a parallel corpus where the special markers are inserted around the  corresponding name entities  in  source and target sentences, with  following  steps.  

\begin{enumerate}[topsep=2pt,itemsep=0pt,partopsep=2pt,parsep=1pt]
\item Detect named entities on the English side of the parallel corpus, using the SpaCy NER system,\footnote{\url{https://spacy.io/models/en\#en_core_web_sm}} which covers 18 types of NER labels. 
\item Translate  the English entity names into the target language, and use string matching to find the corresponding entities from the target sentence. Given a pair of entities in the source and target sentences, we add square brackets (\texttt{[]}) around both of them.
\item Select all sentence pairs that contain more than one \texttt{[]}-marked entity. We also sort the rest of the data based on  length, and include the top-k sentence pairs. In total, we use 5,000 sentence pairs for each language pair.
\end{enumerate}

 We utilize the training  data of  NLLB model\footnote{\url{https://huggingface.co/datasets/allenai/nllb}} as the source of the parallel sentences, and use the sentence pairs  from high-resource language pairs (\textit{en} to \{\textit{de,es,nl,zh,ar}\}), which are selected based on the CoNLL-2002/2003 (\{\textit{de,es,nl}\}) and ACE datasets (\{\textit{zh,ar}\}).



\paragraph{Parameter-efficient Fine-tuning.} To save compute and preserve the translation ability of the model, we  only update the weights in  the last  layer of the encoder and decoder for 200 steps, with a learning rate of 5e-5 and a batch size of 8, which takes around 2 minutes on an A40 GPU. The changes in projection rate and the translation quality during the fine-tuning process are shown in Figure \ref{fig:projection-rate-translation-quality-change}.
The fine-tuned NLLB model is able to improve the projection rate on TyDiQA from 70\% to 96.4\%. TyDiQA is particularly challenging due to its relatively long sentences, and  the translation model sometimes may ignore the inserted markers. 
By fine-tuning on high-resource languages, we found the   model is able to generalize well to the other language pairs. In pilot study, we notice that fine-tuning on low-resource languages, such as African languages in MasakhaNER corpus,  doesn't generalize well and  leads to lower translation quality. We will release all the fine-tuned models.



\paragraph{Fuzzy String Matching.}  To identify the corresponding labels when more than one  projected entity exists in the translation, we design a fuzzy string-matching method.   We first translate  each annotated span in the original sentence independently, resulting in a set of labeled mentions.  To identify labels for the  unlabeled spans in the \texttt{[]}-marked translation, we compare each unlabeled span with the  labeled mentions using the \texttt{ratio()}  function in the \texttt{difflib} library.{\interfootnotelinepenalty100000000\
\footnote{\url{https://docs.python.org/3/library/difflib.html\#difflib.SequenceMatcher.ratio}}}
Two strings are considered matched if they have $>$50\% matched subsequences, and the associated label is assigned to the bracketed span. We also experiment with  matching span labels left-to-right based on their relative position in the text. The results are shown in Table \ref{table:ablation}. Using fuzzy string-matching leads to overall better performance since it can assign the span labels more accurately.



\renewcommand{\arraystretch}{1.2}
\begin{table}[t!]
\centering
\normalsize
\resizebox{1\linewidth}{!}{
\begin{tabular}{ l cc cc}
\toprule
  &  \multicolumn{2}{c}{Original NLLB} & \multicolumn{2}{c}{Fine-tuned NLLB} \\ \cmidrule(lr){2-3} \cmidrule(lr){4-5} 
  & Proj. Rate & F$_1$ & Proj. Rate & F$_1$  \\ 
\midrule
\texttt{[\,]\,}+Fuzzy String Match & 87.0\% & \underline{62.6} & \underline{93.7\%} & \textbf{62.9} \\
\texttt{[\,]\,}+Match by Sequence & 87.7\% & \underline{62.6} & \textbf{94.4\%} & 62.3\\
\bottomrule
\end{tabular}
}
\vspace{-7pt}
\caption{A  comparison of varied methods to translate sentences and assign labels on  the devset of MasakhaNER 2.0 corpus. ``Proj.Rate'' denotes the projection rate, which is defined in $\S$\ref{sec:choice-of-markers}.} 
\label{table:ablation}
\vspace{-18pt}
\end{table}

Putting all the improvements together,  we call this improved variant of marker-based method   \textbf{\textsc{EasyProject}} for \textbf{eas}il\textbf{y} \textbf{project}ing labels.  

\section{Experiments}
In this section, we comprehensively evaluate the effectiveness of {\sc EasyProject} and analyze the key factors that impact the performance of  cross-lingual transfer learning. 

\subsection{Comparison to Alignment-based Method}
\label{sec:comparsion_to_alignment}

We first compare EasyProject with  the traditional pipeline approach based on state-of-the-art bilingual word alignment models. For both methods, we apply  a simple filtering rule that removes sentences with different numbers of annotations before and after projection.

\vspace{-2.5pt}
\paragraph{Bilingual Word Alignment.} We experiment with two state-of-the-art neural word aligners: (1) the unsupervised version of \textbf{Awesome-align} \cite{dou2021awesome} and its supervised version, which we extended from 5 to 39 languages for this paper, and (2) \textbf{QA-align} \cite{nagata2020supervised} which formulates the word alignment problem as SQuAD-style QA task. More details on the 
 word alignment models  can be found in Appendix \ref{appendix:word_alignment}.


\renewcommand{\arraystretch}{1.1}
\begin{table}[b!]
\centering
\small
\vspace{-10pt}
\resizebox{0.98\linewidth}{!}{
\begin{tabular}{ccccc}
\toprule
\multicolumn{2}{c}{Fine-tune$_\textit{en}$ } & NLLB+Aligner &  \multicolumn{2}{c}{NLLB+Markers}  \\ \cmidrule(lr){1-2} \cmidrule(lr){3-3} \cmidrule(lr){4-5} 
Ref. & mDeBERTa & Awesome-align & XML & EasyProject. \\

\midrule
56.9 & 55.0 & 63.2 (+8.2) & 63.8 (+8.8) & \textbf{64.3} (+9.3) \\

\bottomrule
\end{tabular}
}
\vspace{-5pt}
\caption{Average results over 18 African languages on the MasaKhaNER 2.0 corpus, as two languages are not supported by NLLB model.   The mDeBERTa model is used here sinces it has   the strongest performance on African languages in  the original paper \cite{adelani2022masakhaner}, where the ``Ref'' column  is from. Full results on each languages are in Table \ref{table:results-on-masakha2} in Appendix \ref{sec:results-on-masakha2}
.}
\vspace{-5pt}
\label{table:masakha}
\end{table}

\renewcommand{\arraystretch}{1.05}
\begin{table}[b!]
\centering
\small
\begin{tabular}{lccc}
\toprule
Model & Ref. & Fine-tune$_{\textit{en}}$ & GMT+EasyProject\\
\midrule
mT5$_{\text{large}}$ & 58.2 & 61.2 & \textbf{68.5} (+7.3)\\
mT5$_{\text{XL}}$ & 65.1 & 62.9 & \textbf{68.6} (+5.7)\\

XLM-R$_{\text{large}}$ & 63.3 & 64.3 & \textbf{68.9} (+4.6)\\
\bottomrule
\end{tabular}
\vspace{-5pt}
\caption{Average results  for  mT5 models over 39 target languages  on WikiANN. The XLM-R model is listed for comparison. ``Ref.'' column is the performance from prior work (\citeauthor{he2021effectiveness, xue2021mt5}).  }
 \vspace{-5pt}
\label{table:mt5}
\end{table}

\begin{table*}[phbt!]
\centering
\small
\vspace{-5pt}
\resizebox{0.95\textwidth}{!}{
\renewcommand{\arraystretch}{0.75}
\begin{tabularx}{\textwidth}{p{8mm}c*{15}{Y}}
\toprule
\multicolumn{2}{c}{\multirow{2}{*}{\textit{en} $\rightarrow$ Lang.}}  & \multicolumn{2}{c}{Fine-tune$_{\text{en}}$} & \multicolumn{3}{c}{NLLB+Word Align} & \multicolumn{3}{c}{NLLB+Markers} & \multicolumn{3}{c}{GMT+Word Align} & \multicolumn{3}{c}{GMT+Markers}\\
\cmidrule(lr){3-4} \cmidrule(lr){5-7} \cmidrule(lr){8-10} \cmidrule(lr){11-13} \cmidrule(lr){14-16} 

& & {\scriptsize Ref.} & {\scriptsize XLM$_R$} & {\scriptsize QAali.} & {\scriptsize Awes.} & \multicolumn{1}{r}{{\scriptsize Awes$_{\text{ft}}$}} & {\scriptsize XML } & \multicolumn{2}{c}{\scriptsize {EProj.} ($\Delta_{\text{XLM}_R}$)}  & {\scriptsize QAali.} & {\scriptsize Awes.} & \multicolumn{1}{r}{{\scriptsize Awes$_{\text{ft}}$}} & {\scriptsize XML}  & \multicolumn{2}{c}{\scriptsize{EProj.} ($\Delta_{\text{XLM}_R})$}\\
\midrule
\multirow{40}{*}{NER} 
&\textit{yo}&41.3&37.1&-&\cellcolor[HTML]{52BE80}73.2&\cellcolor[HTML]{52BE80}78.0&\cellcolor[HTML]{52BE80}68.7&\multicolumn{2}{l}{\cellcolor[HTML]{52BE80} 77.7 (+40.6)}&-&\cellcolor[HTML]{52BE80}72.1&\cellcolor[HTML]{52BE80}66.1&\cellcolor[HTML]{52BE80}71.8&\multicolumn{2}{l}{\cellcolor[HTML]{52BE80} 73.8 (+36.7)}\\
&\textit{ja}&18.3&18.0&\cellcolor[HTML]{A9DFBF}19.3&\cellcolor[HTML]{7DCEA0}23.4&\cellcolor[HTML]{A9DFBF}22.4&\cellcolor[HTML]{FDEDEC}17.3&\multicolumn{2}{l}{\cellcolor[HTML]{52BE80} 45.5 (+27.5)}&\cellcolor[HTML]{A9DFBF}19.3&\cellcolor[HTML]{7DCEA0}23.0&\cellcolor[HTML]{A9DFBF}22.6&\cellcolor[HTML]{52BE80}42.0&\multicolumn{2}{l}{\cellcolor[HTML]{52BE80} 43.5 (+25.5)}\\
&\textit{zh}&25.8&27.1&\cellcolor[HTML]{52BE80}47.6&\cellcolor[HTML]{7DCEA0}36.0&\cellcolor[HTML]{7DCEA0}34.0&\cellcolor[HTML]{52BE80}46.2&\multicolumn{2}{l}{\cellcolor[HTML]{52BE80} 46.6 (+19.5)}&\cellcolor[HTML]{52BE80}45.2&\cellcolor[HTML]{52BE80}43.3&\cellcolor[HTML]{52BE80}39.6&\cellcolor[HTML]{52BE80}43.8&\multicolumn{2}{l}{\cellcolor[HTML]{52BE80} 45.9 (+18.8)}\\
&\textit{th}&1.5&0.7&-&\cellcolor[HTML]{A9DFBF}2.6&\cellcolor[HTML]{A9DFBF}2.5&\cellcolor[HTML]{7DCEA0}8.8&\multicolumn{2}{l}{\cellcolor[HTML]{52BE80} 14.0 (+13.3)}&-&\cellcolor[HTML]{E9F7EF}1.2&\cellcolor[HTML]{E9F7EF}1.3&\cellcolor[HTML]{52BE80}14.7&\multicolumn{2}{l}{\cellcolor[HTML]{52BE80} 15.1 (+14.4)}\\
&\textit{ur}&54.2&63.6&-&\cellcolor[HTML]{7DCEA0}71.6&\cellcolor[HTML]{7DCEA0}71.8&\cellcolor[HTML]{52BE80}74.4&\multicolumn{2}{l}{\cellcolor[HTML]{52BE80} 74.7 (+11.1)}&-&\cellcolor[HTML]{7DCEA0}70.2&\cellcolor[HTML]{7DCEA0}72.3&\cellcolor[HTML]{52BE80}76.3&\multicolumn{2}{l}{\cellcolor[HTML]{52BE80} 74.7 (+11.1)}\\
&\textit{he}&54.1&56.0&-&\cellcolor[HTML]{A9DFBF}58.3&\cellcolor[HTML]{A9DFBF}58.1&\cellcolor[HTML]{7DCEA0}61.1&\multicolumn{2}{l}{\cellcolor[HTML]{7DCEA0} 63.4 (+7.4)}&-&\cellcolor[HTML]{A9DFBF}59.6&\cellcolor[HTML]{A9DFBF}60.2&\cellcolor[HTML]{7DCEA0}63.7&\multicolumn{2}{l}{\cellcolor[HTML]{52BE80} 67.1 (+11.1)}\\
&\textit{ms}&69.8&64.1&-&\cellcolor[HTML]{7DCEA0}69.4&\cellcolor[HTML]{7DCEA0}72.7&\cellcolor[HTML]{52BE80}74.6&\multicolumn{2}{l}{\cellcolor[HTML]{7DCEA0} 73.9 (+9.8)}&-&\cellcolor[HTML]{7DCEA0}73.0&\cellcolor[HTML]{7DCEA0}73.8&\cellcolor[HTML]{7DCEA0}73.2&\multicolumn{2}{l}{\cellcolor[HTML]{52BE80} 74.1 (+10.0)}\\
&\textit{my}&51.3&53.5&-&\cellcolor[HTML]{7DCEA0}61.6&\cellcolor[HTML]{7DCEA0}62.9&\cellcolor[HTML]{A9DFBF}56.2&\multicolumn{2}{l}{\cellcolor[HTML]{7DCEA0} 60.1 (+6.6)}&-&\cellcolor[HTML]{7DCEA0}60.2&\cellcolor[HTML]{7DCEA0}60.1&\cellcolor[HTML]{A9DFBF}57.0&\multicolumn{2}{l}{\cellcolor[HTML]{7DCEA0} 62.0 (+8.5)}\\
&\textit{ar}&43.7&48.5&\cellcolor[HTML]{E9F7EF}49.3&\cellcolor[HTML]{E9F7EF}48.7&\cellcolor[HTML]{FDEDEC}47.6&\cellcolor[HTML]{FADBD8}45.9&\multicolumn{2}{l}{\cellcolor[HTML]{A9DFBF} 50.5 (+2.0)}&\cellcolor[HTML]{A9DFBF}50.7&\cellcolor[HTML]{A9DFBF}50.9&\cellcolor[HTML]{A9DFBF}51.2&\cellcolor[HTML]{A9DFBF}51.3&\multicolumn{2}{l}{\cellcolor[HTML]{7DCEA0} 56.3 (+7.8)}\\
&\textit{jv}&58.4&62.3&-&\cellcolor[HTML]{A9DFBF}64.8&\cellcolor[HTML]{FDEDEC}61.6&\cellcolor[HTML]{7DCEA0}67.7&\multicolumn{2}{l}{\cellcolor[HTML]{A9DFBF} 67.0 (+4.7)}&-&\cellcolor[HTML]{A9DFBF}64.6&\cellcolor[HTML]{7DCEA0}68.8&\cellcolor[HTML]{7DCEA0}69.2&\multicolumn{2}{l}{\cellcolor[HTML]{7DCEA0} 69.8 (+7.5)}\\
&\textit{tl}&72.2&73.0&-&\cellcolor[HTML]{7DCEA0}80.1&\cellcolor[HTML]{7DCEA0}78.8&\cellcolor[HTML]{7DCEA0}79.5&\multicolumn{2}{l}{\cellcolor[HTML]{7DCEA0} 79.3 (+6.3)}&-&\cellcolor[HTML]{7DCEA0}80.4&\cellcolor[HTML]{7DCEA0}80.4&\cellcolor[HTML]{7DCEA0}79.9&\multicolumn{2}{l}{\cellcolor[HTML]{7DCEA0} 80.0 (+7.0)}\\
&\textit{hi}&71.0&69.5&-&\cellcolor[HTML]{A9DFBF}73.9&\cellcolor[HTML]{A9DFBF}73.4&\cellcolor[HTML]{A9DFBF}73.8&\multicolumn{2}{l}{\cellcolor[HTML]{A9DFBF} 74.4 (+4.9)}&-&\cellcolor[HTML]{7DCEA0}75.6&\cellcolor[HTML]{7DCEA0}76.0&\cellcolor[HTML]{7DCEA0}75.9&\multicolumn{2}{l}{\cellcolor[HTML]{7DCEA0} 75.7 (+6.2)}\\
&\textit{ka}&68.9&68.8&-&\cellcolor[HTML]{7DCEA0}74.5&\cellcolor[HTML]{7DCEA0}75.0&\cellcolor[HTML]{A9DFBF}70.4&\multicolumn{2}{l}{\cellcolor[HTML]{7DCEA0} 74.2 (+5.4)}&-&\cellcolor[HTML]{A9DFBF}73.5&\cellcolor[HTML]{A9DFBF}73.2&\cellcolor[HTML]{A9DFBF}72.7&\multicolumn{2}{l}{\cellcolor[HTML]{7DCEA0} 74.7 (+5.9)}\\
&\textit{bn}&76.3&75.1&-&\cellcolor[HTML]{7DCEA0}80.5&\cellcolor[HTML]{7DCEA0}80.2&\cellcolor[HTML]{7DCEA0}80.1&\multicolumn{2}{l}{\cellcolor[HTML]{7DCEA0} 80.7 (+5.6)}&-&\cellcolor[HTML]{7DCEA0}82.0&\cellcolor[HTML]{7DCEA0}81.7&\cellcolor[HTML]{7DCEA0}80.6&\multicolumn{2}{l}{\cellcolor[HTML]{7DCEA0} 80.9 (+5.8)}\\
&\textit{ta}&56.9&58.8&-&\cellcolor[HTML]{A9DFBF}63.1&\cellcolor[HTML]{A9DFBF}63.7&\cellcolor[HTML]{F5B7B1}53.8&\multicolumn{2}{l}{\cellcolor[HTML]{A9DFBF} 63.5 (+4.7)}&-&\cellcolor[HTML]{A9DFBF}62.4&\cellcolor[HTML]{A9DFBF}63.2&\cellcolor[HTML]{7DCEA0}63.9&\multicolumn{2}{l}{\cellcolor[HTML]{7DCEA0} 64.3 (+5.5)}\\
&\textit{eu}&62.1&63.6&-&\cellcolor[HTML]{7DCEA0}70.3&\cellcolor[HTML]{7DCEA0}70.0&\cellcolor[HTML]{A9DFBF}64.7&\multicolumn{2}{l}{\cellcolor[HTML]{7DCEA0} 68.7 (+5.1)}&-&\cellcolor[HTML]{7DCEA0}69.8&\cellcolor[HTML]{A9DFBF}66.5&\cellcolor[HTML]{A9DFBF}67.5&\multicolumn{2}{l}{\cellcolor[HTML]{7DCEA0} 69.0 (+5.4)}\\
&\textit{ko}&58.0&57.9&-&\cellcolor[HTML]{A9DFBF}61.1&\cellcolor[HTML]{A9DFBF}60.6&\cellcolor[HTML]{A9DFBF}59.4&\multicolumn{2}{l}{\cellcolor[HTML]{E9F7EF} 58.0 (+0.1)}&-&\cellcolor[HTML]{7DCEA0}62.9&\cellcolor[HTML]{A9DFBF}62.4&\cellcolor[HTML]{A9DFBF}61.7&\multicolumn{2}{l}{\cellcolor[HTML]{A9DFBF} 61.9 (+4.0)}\\
&\textit{mr}&64.1&63.9&-&\cellcolor[HTML]{FDEDEC}63.6&\cellcolor[HTML]{E9F7EF}64.0&\cellcolor[HTML]{FADBD8}62.9&\multicolumn{2}{l}{\cellcolor[HTML]{A9DFBF} 64.9 (+1.0)}&-&\cellcolor[HTML]{FADBD8}62.6&\cellcolor[HTML]{FADBD8}61.2&\cellcolor[HTML]{E9F7EF}64.0&\multicolumn{2}{l}{\cellcolor[HTML]{A9DFBF} 67.1 (+3.2)}\\
&\textit{sw}&70.0&68.5&-&\cellcolor[HTML]{A9DFBF}70.6&\cellcolor[HTML]{A9DFBF}71.5&\cellcolor[HTML]{A9DFBF}70.1&\multicolumn{2}{l}{\cellcolor[HTML]{A9DFBF} 69.7 (+1.2)}&-&\cellcolor[HTML]{A9DFBF}70.2&\cellcolor[HTML]{A9DFBF}71.5&\cellcolor[HTML]{A9DFBF}72.2&\multicolumn{2}{l}{\cellcolor[HTML]{A9DFBF} 70.7 (+2.2)}\\
&\textit{vi}&77.2&74.2&-&\cellcolor[HTML]{FADBD8}70.4&\cellcolor[HTML]{F5B7B1}65.8&\cellcolor[HTML]{A9DFBF}77.8&\multicolumn{2}{l}{\cellcolor[HTML]{A9DFBF} 77.5 (+3.3)}&-&\cellcolor[HTML]{FADBD8}70.4&\cellcolor[HTML]{F5B7B1}67.2&\cellcolor[HTML]{A9DFBF}77.5&\multicolumn{2}{l}{\cellcolor[HTML]{A9DFBF} 76.0 (+1.8)}\\
&\textit{te}&52.3&55.6&-&\cellcolor[HTML]{A9DFBF}57.7&\cellcolor[HTML]{E9F7EF}56.3&\cellcolor[HTML]{FADBD8}51.8&\multicolumn{2}{l}{\cellcolor[HTML]{E9F7EF} 55.9 (+0.3)}&-&\cellcolor[HTML]{A9DFBF}57.4&\cellcolor[HTML]{A9DFBF}56.8&\cellcolor[HTML]{A9DFBF}57.6&\multicolumn{2}{l}{\cellcolor[HTML]{A9DFBF} 57.4 (+1.8)}\\
&\textit{id}&52.3&52.4&-&\cellcolor[HTML]{A9DFBF}53.5&\cellcolor[HTML]{A9DFBF}55.3&\cellcolor[HTML]{E9F7EF}52.7&\multicolumn{2}{l}{\cellcolor[HTML]{E9F7EF} 53.1 (+0.7)}&-&\cellcolor[HTML]{E9F7EF}52.7&\cellcolor[HTML]{A9DFBF}55.0&\cellcolor[HTML]{A9DFBF}57.3&\multicolumn{2}{l}{\cellcolor[HTML]{A9DFBF} 53.9 (+1.5)}\\
&\textit{ml}&65.8&63.5&-&\cellcolor[HTML]{FDEDEC}63.2&\cellcolor[HTML]{A9DFBF}64.8&\cellcolor[HTML]{F5B7B1}56.5&\multicolumn{2}{l}{\cellcolor[HTML]{FADBD8} 61.3 (-2.2)}&-&\cellcolor[HTML]{FADBD8}61.9&\cellcolor[HTML]{FDEDEC}63.0&\cellcolor[HTML]{A9DFBF}68.1&\multicolumn{2}{l}{\cellcolor[HTML]{E9F7EF} 64.3 (+0.8)}\\
&\textit{es}&68.8&74.8&-&\cellcolor[HTML]{FADBD8}72.2&\cellcolor[HTML]{FADBD8}70.2&\cellcolor[HTML]{FADBD8}73.3&\multicolumn{2}{l}{\cellcolor[HTML]{FADBD8} 71.7 (-3.1)}&-&\cellcolor[HTML]{FADBD8}71.3&\cellcolor[HTML]{FADBD8}72.6&\cellcolor[HTML]{FADBD8}73.5&\multicolumn{2}{l}{\cellcolor[HTML]{E9F7EF} 75.6 (+0.8)}\\
&\textit{de}&77.9&79.4&\cellcolor[HTML]{E9F7EF}79.7&\cellcolor[HTML]{E9F7EF}79.5&\cellcolor[HTML]{E9F7EF}79.6&\cellcolor[HTML]{A9DFBF}81.5&\multicolumn{2}{l}{\cellcolor[HTML]{E9F7EF} 80.0 (+0.6)}&\cellcolor[HTML]{E9F7EF}79.5&\cellcolor[HTML]{E9F7EF}80.0&\cellcolor[HTML]{E9F7EF}79.4&\cellcolor[HTML]{E9F7EF}79.8&\multicolumn{2}{l}{\cellcolor[HTML]{E9F7EF} 80.2 (+0.8)}\\
&\textit{kk}&49.8&53.5&-&\cellcolor[HTML]{E9F7EF}53.5&\cellcolor[HTML]{E9F7EF}53.9&\cellcolor[HTML]{F1948A}40.4&\multicolumn{2}{l}{\cellcolor[HTML]{E9F7EF} 54.0 (+0.5)}&-&\cellcolor[HTML]{FDEDEC}53.2&\cellcolor[HTML]{A9DFBF}55.1&\cellcolor[HTML]{FADBD8}51.3&\multicolumn{2}{l}{\cellcolor[HTML]{E9F7EF} 54.2 (+0.7)}\\
&\textit{fr}&79.0&80.1&\cellcolor[HTML]{E9F7EF}80.7&\cellcolor[HTML]{FDEDEC}79.8&\cellcolor[HTML]{E9F7EF}80.9&\cellcolor[HTML]{E9F7EF}80.9&\multicolumn{2}{l}{\cellcolor[HTML]{A9DFBF} 81.5 (+1.4)}&\cellcolor[HTML]{FDEDEC}79.6&\cellcolor[HTML]{E9F7EF}80.7&\cellcolor[HTML]{FDEDEC}79.4&\cellcolor[HTML]{A9DFBF}81.5&\multicolumn{2}{l}{\cellcolor[HTML]{E9F7EF} 80.8 (+0.7)}\\
&\textit{af}&77.6&78.6&-&\cellcolor[HTML]{E9F7EF}79.3&\cellcolor[HTML]{FDEDEC}78.4&\cellcolor[HTML]{E9F7EF}79.1&\multicolumn{2}{l}{\cellcolor[HTML]{E9F7EF} 79.4 (+0.8)}&-&\cellcolor[HTML]{E9F7EF}79.1&\cellcolor[HTML]{E9F7EF}78.9&\cellcolor[HTML]{E9F7EF}79.0&\multicolumn{2}{l}{\cellcolor[HTML]{E9F7EF} 79.2 (+0.6)}\\
&\textit{et}&78.0&79.6&-&\cellcolor[HTML]{A9DFBF}80.7&\cellcolor[HTML]{FDEDEC}79.2&\cellcolor[HTML]{E9F7EF}80.2&\multicolumn{2}{l}{\cellcolor[HTML]{E9F7EF} 79.9 (+0.3)}&-&\cellcolor[HTML]{E9F7EF}80.2&\cellcolor[HTML]{E9F7EF}79.6&\cellcolor[HTML]{FADBD8}78.6&\multicolumn{2}{l}{\cellcolor[HTML]{E9F7EF} 80.1 (+0.5)}\\
&\textit{hu}&79.3&81.0&-&\cellcolor[HTML]{FDEDEC}80.3&\cellcolor[HTML]{FADBD8}79.8&\cellcolor[HTML]{FADBD8}79.7&\multicolumn{2}{l}{\cellcolor[HTML]{FDEDEC} 80.4 (-0.6)}&-&\cellcolor[HTML]{FADBD8}79.9&\cellcolor[HTML]{FADBD8}79.7&\cellcolor[HTML]{FDEDEC}80.6&\multicolumn{2}{l}{\cellcolor[HTML]{FDEDEC} 80.7 (-0.3)}\\
&\textit{fi}&78.6&80.6&-&\cellcolor[HTML]{E9F7EF}81.0&\cellcolor[HTML]{E9F7EF}80.9&\cellcolor[HTML]{FDEDEC}80.4&\multicolumn{2}{l}{\cellcolor[HTML]{FDEDEC} 79.8 (-0.8)}&-&\cellcolor[HTML]{E9F7EF}80.7&\cellcolor[HTML]{FDEDEC}79.7&\cellcolor[HTML]{FADBD8}78.8&\multicolumn{2}{l}{\cellcolor[HTML]{FDEDEC} 80.3 (-0.3)}\\
&\textit{it}&81.1&81.3&-&\cellcolor[HTML]{FDEDEC}80.5&\cellcolor[HTML]{FDEDEC}80.5&\cellcolor[HTML]{E9F7EF}81.9&\multicolumn{2}{l}{\cellcolor[HTML]{FDEDEC} 81.2 (-0.1)}&-&\cellcolor[HTML]{FADBD8}80.3&\cellcolor[HTML]{FDEDEC}80.4&\cellcolor[HTML]{FDEDEC}81.1&\multicolumn{2}{l}{\cellcolor[HTML]{FDEDEC} 80.9 (-0.4)}\\
&\textit{tr}&78.9&80.3&-&\cellcolor[HTML]{E9F7EF}80.6&\cellcolor[HTML]{E9F7EF}81.0&\cellcolor[HTML]{FDEDEC}80.1&\multicolumn{2}{l}{\cellcolor[HTML]{FDEDEC} 79.5 (-0.8)}&-&\cellcolor[HTML]{FDEDEC}80.1&\cellcolor[HTML]{FDEDEC}80.2&\cellcolor[HTML]{A9DFBF}81.5&\multicolumn{2}{l}{\cellcolor[HTML]{FDEDEC} 79.6 (-0.7)}\\
&\textit{nl}&84.3&84.1&-&\cellcolor[HTML]{FDEDEC}83.4&\cellcolor[HTML]{FDEDEC}83.3&\cellcolor[HTML]{FADBD8}83.0&\multicolumn{2}{l}{\cellcolor[HTML]{FDEDEC} 83.4 (-0.7)}&-&\cellcolor[HTML]{FDEDEC}83.5&\cellcolor[HTML]{FADBD8}82.9&\cellcolor[HTML]{FADBD8}83.0&\multicolumn{2}{l}{\cellcolor[HTML]{FADBD8} 83.1 (-1.0)}\\
&\textit{bg}&81.2&82.1&-&\cellcolor[HTML]{FADBD8}80.2&\cellcolor[HTML]{FADBD8}78.8&\cellcolor[HTML]{FDEDEC}81.9&\multicolumn{2}{l}{\cellcolor[HTML]{E9F7EF} 82.5 (+0.4)}&-&\cellcolor[HTML]{FADBD8}80.9&\cellcolor[HTML]{FADBD8}79.7&\cellcolor[HTML]{E9F7EF}82.5&\multicolumn{2}{l}{\cellcolor[HTML]{FADBD8} 80.6 (-1.5)}\\
&\textit{pt}&79.6&82.0&-&\cellcolor[HTML]{FADBD8}80.9&\cellcolor[HTML]{FADBD8}80.4&\cellcolor[HTML]{E9F7EF}82.6&\multicolumn{2}{l}{\cellcolor[HTML]{FDEDEC} 81.9 (-0.1)}&-&\cellcolor[HTML]{FADBD8}79.0&\cellcolor[HTML]{FADBD8}80.2&\cellcolor[HTML]{FADBD8}80.6&\multicolumn{2}{l}{\cellcolor[HTML]{FADBD8} 80.1 (-1.9)}\\
&\textit{ru}&71.5&71.1&-&\cellcolor[HTML]{FADBD8}68.9&\cellcolor[HTML]{FADBD8}68.1&\cellcolor[HTML]{FADBD8}70.0&\multicolumn{2}{l}{\cellcolor[HTML]{FDEDEC} 70.3 (-0.8)}&-&\cellcolor[HTML]{FADBD8}67.4&\cellcolor[HTML]{FADBD8}66.8&\cellcolor[HTML]{FADBD8}67.4&\multicolumn{2}{l}{\cellcolor[HTML]{FADBD8} 68.2 (-2.9)}\\
&\textit{el}&77.2&79.3&-&\cellcolor[HTML]{FADBD8}76.3&\cellcolor[HTML]{FADBD8}75.7&\cellcolor[HTML]{FADBD8}77.7&\multicolumn{2}{l}{\cellcolor[HTML]{F5B7B1} 74.1 (-5.2)}&-&\cellcolor[HTML]{F5B7B1}73.1&\cellcolor[HTML]{FADBD8}75.2&\cellcolor[HTML]{FADBD8}76.2&\multicolumn{2}{l}{\cellcolor[HTML]{FADBD8} 75.0 (-4.3)}\\
&\textit{fa}&61.1&64.3&-&\cellcolor[HTML]{F1948A}41.5&\cellcolor[HTML]{F1948A}47.3&\cellcolor[HTML]{F1948A}51.3&\multicolumn{2}{l}{\cellcolor[HTML]{F1948A} 52.1 (-12.2)}&-&\cellcolor[HTML]{F1948A}52.9&\cellcolor[HTML]{F1948A}52.4&\cellcolor[HTML]{F1948A}45.5&\multicolumn{2}{l}{\cellcolor[HTML]{F1948A} 52.0 (-12.3)}\\
\cmidrule(lr){2-16}
&\textit{AVG}&63.3&64.3&-&\cellcolor[HTML]{A9DFBF}66.4&\cellcolor[HTML]{A9DFBF}66.4&\cellcolor[HTML]{A9DFBF}66.1&\multicolumn{2}{l}{\cellcolor[HTML]{A9DFBF} 68.4 (+4.1)}&-&\cellcolor[HTML]{A9DFBF}66.7&\cellcolor[HTML]{A9DFBF}66.6&\cellcolor[HTML]{A9DFBF}68.3&\multicolumn{2}{l}{\cellcolor[HTML]{A9DFBF} 68.9 (+4.6)}\\
\midrule
\multirow{10}{*}{QA} 
&\textit{ko}&31.9&56.1&-&\cellcolor[HTML]{F1948A}36.9&\cellcolor[HTML]{F1948A}36.4&\cellcolor[HTML]{7DCEA0}64.8&\multicolumn{2}{l}{\cellcolor[HTML]{52BE80} 67.7 (+11.6)}&-&\cellcolor[HTML]{F1948A}37.6&\cellcolor[HTML]{F1948A}37.1&\cellcolor[HTML]{A9DFBF}60.9&\multicolumn{2}{l}{\cellcolor[HTML]{7DCEA0} 65.0 (+8.9)}\\
&\textit{bn}&64.0&66.0&-&\cellcolor[HTML]{7DCEA0}71.1&\cellcolor[HTML]{7DCEA0}72.6&\cellcolor[HTML]{FADBD8}63.7&\multicolumn{2}{l}{\cellcolor[HTML]{A9DFBF} 69.6 (+3.6)}&-&\cellcolor[HTML]{7DCEA0}73.6&\cellcolor[HTML]{A9DFBF}69.3&\cellcolor[HTML]{7DCEA0}74.4&\multicolumn{2}{l}{\cellcolor[HTML]{7DCEA0} 71.0 (+5.0)}\\
&\textit{fi}&70.5&69.7&-&\cellcolor[HTML]{7DCEA0}74.9&\cellcolor[HTML]{A9DFBF}74.0&\cellcolor[HTML]{A9DFBF}73.0&\multicolumn{2}{l}{\cellcolor[HTML]{A9DFBF} 73.3 (+3.6)}&-&\cellcolor[HTML]{7DCEA0}74.9&\cellcolor[HTML]{7DCEA0}74.9&\cellcolor[HTML]{A9DFBF}73.1&\multicolumn{2}{l}{\cellcolor[HTML]{A9DFBF} 74.0 (+4.3)}\\
&\textit{te}&70.1&72.9&-&\cellcolor[HTML]{A9DFBF}74.9&\cellcolor[HTML]{A9DFBF}74.6&\cellcolor[HTML]{FADBD8}69.9&\multicolumn{2}{l}{\cellcolor[HTML]{7DCEA0} 78.3 (+5.4)}&-&\cellcolor[HTML]{A9DFBF}75.9&\cellcolor[HTML]{FADBD8}69.9&\cellcolor[HTML]{A9DFBF}77.0&\multicolumn{2}{l}{\cellcolor[HTML]{A9DFBF} 77.0 (+4.1)}\\
&\textit{ar}&67.6&72.4&\cellcolor[HTML]{A9DFBF}74.2&\cellcolor[HTML]{A9DFBF}76.8&\cellcolor[HTML]{A9DFBF}76.4&\cellcolor[HTML]{E9F7EF}72.7&\multicolumn{2}{l}{\cellcolor[HTML]{A9DFBF} 75.9 (+3.5)}&\cellcolor[HTML]{A9DFBF}74.0&\cellcolor[HTML]{A9DFBF}76.3&\cellcolor[HTML]{A9DFBF}76.6&\cellcolor[HTML]{A9DFBF}75.8&\multicolumn{2}{l}{\cellcolor[HTML]{A9DFBF} 76.4 (+4.0)}\\
&\textit{sw}&66.1&69.9&-&\cellcolor[HTML]{A9DFBF}73.0&\cellcolor[HTML]{A9DFBF}74.7&\cellcolor[HTML]{A9DFBF}72.4&\multicolumn{2}{l}{\cellcolor[HTML]{A9DFBF} 73.4 (+3.5)}&-&\cellcolor[HTML]{A9DFBF}72.3&\cellcolor[HTML]{A9DFBF}73.4&\cellcolor[HTML]{A9DFBF}73.6&\multicolumn{2}{l}{\cellcolor[HTML]{A9DFBF} 73.5 (+3.6)}\\
&\textit{ru}&67.0&66.5&-&\cellcolor[HTML]{A9DFBF}70.9&\cellcolor[HTML]{7DCEA0}71.5&\cellcolor[HTML]{A9DFBF}69.1&\multicolumn{2}{l}{\cellcolor[HTML]{A9DFBF} 70.4 (+3.9)}&-&\cellcolor[HTML]{7DCEA0}71.6&\cellcolor[HTML]{A9DFBF}69.7&\cellcolor[HTML]{A9DFBF}70.2&\multicolumn{2}{l}{\cellcolor[HTML]{A9DFBF} 69.8 (+3.3)}\\
&\textit{id}&77.4&78.0&-&\cellcolor[HTML]{A9DFBF}81.6&\cellcolor[HTML]{A9DFBF}81.1&\cellcolor[HTML]{A9DFBF}79.6&\multicolumn{2}{l}{\cellcolor[HTML]{A9DFBF} 80.3 (+2.3)}&-&\cellcolor[HTML]{A9DFBF}80.4&\cellcolor[HTML]{A9DFBF}81.3&\cellcolor[HTML]{E9F7EF}78.9&\multicolumn{2}{l}{\cellcolor[HTML]{A9DFBF} 79.7 (+1.7)}\\
\cmidrule(lr){2-16}
&\textit{AVG}&64.3&68.9&-&\cellcolor[HTML]{A9DFBF}70.0&\cellcolor[HTML]{A9DFBF}70.2&\cellcolor[HTML]{A9DFBF}70.7&\multicolumn{2}{l}{\cellcolor[HTML]{A9DFBF} 73.6 (+4.7)}&-&\cellcolor[HTML]{A9DFBF}70.3&\cellcolor[HTML]{E9F7EF}69.0&\cellcolor[HTML]{A9DFBF}73.0&\multicolumn{2}{l}{\cellcolor[HTML]{A9DFBF} 73.3 (+4.4)}\\
\bottomrule
\end{tabularx}}

\resizebox{0.95\textwidth}{!}{
\renewcommand{\arraystretch}{0.75}
\begin{tabularx}{\textwidth}{p{6.5mm}p{8.6mm}c*{12}{Y}}
\toprule
\multirow{3}{*}{Event Extraction}  & & Fine-tune$_{\text{en}}$ & \multicolumn{3}{c}{NLLB+Word Align} & \multicolumn{3}{c}{NLLB+Markers} & \multicolumn{3}{c}{GMT+Word Align} & \multicolumn{3}{c}{GMT+Markers}\\
\cmidrule(lr){3-3} \cmidrule(lr){4-6} \cmidrule(lr){7-9} \cmidrule(lr){10-12} \cmidrule(lr){13-15} 

&  & XLM$_R$ & {\scriptsize QAali.} & {\scriptsize Awes.} & \multicolumn{1}{r}{{\scriptsize Awes$_{\text{ft}}$}} & {\scriptsize XML } & \multicolumn{2}{c}{\scriptsize {EProj.} ($\Delta_{\text{XLM}_R})$}  & {\scriptsize QAali.} & {\scriptsize Awes.} & \multicolumn{1}{r}{{\scriptsize Awes$_{\text{ft}}$}} & {\scriptsize XML}  & \multicolumn{2}{c}{\scriptsize{EProj.} ($\Delta_{\text{XLM}_R})$}\\
\midrule
\multirow{7}{*}{\textit{Arabic}}
&\textit{Entity}&69.2&\cellcolor[HTML]{A9DFBF}74.1&\cellcolor[HTML]{7DCEA0}74.2&\cellcolor[HTML]{7DCEA0}74.2&\cellcolor[HTML]{A9DFBF}73.6&\multicolumn{2}{l}{\cellcolor[HTML]{A9DFBF} 73.8 (+4.6)}&\cellcolor[HTML]{7DCEA0}74.4&\cellcolor[HTML]{7DCEA0}74.3&\cellcolor[HTML]{A9DFBF}74.0&\cellcolor[HTML]{A9DFBF}73.7&\multicolumn{2}{l}{\cellcolor[HTML]{A9DFBF} 74.0 (+4.8)}\\
&\textit{Relation}&28.1&\cellcolor[HTML]{7DCEA0}34.7&\cellcolor[HTML]{7DCEA0}35.2&\cellcolor[HTML]{A9DFBF}30.8&\cellcolor[HTML]{A9DFBF}30.8&\multicolumn{2}{l}{\cellcolor[HTML]{A9DFBF} 30.7 (+2.6)}&\cellcolor[HTML]{7DCEA0}34.8&\cellcolor[HTML]{7DCEA0}33.1&\cellcolor[HTML]{7DCEA0}34.2&\cellcolor[HTML]{A9DFBF}31.8&\multicolumn{2}{l}{\cellcolor[HTML]{7DCEA0} 33.7 (+5.6)}\\
&\textit{Trig-I}&42.7&\cellcolor[HTML]{E9F7EF}43.5&\cellcolor[HTML]{E9F7EF}43.0&\cellcolor[HTML]{A9DFBF}44.7&\cellcolor[HTML]{E9F7EF}43.3&\multicolumn{2}{l}{\cellcolor[HTML]{A9DFBF} 43.7 (+1.0)}&\cellcolor[HTML]{E9F7EF}43.6&\cellcolor[HTML]{A9DFBF}44.2&\cellcolor[HTML]{A9DFBF}43.7&\cellcolor[HTML]{A9DFBF}43.8&\multicolumn{2}{l}{\cellcolor[HTML]{A9DFBF} 44.0 (+1.3)}\\
&\textit{Trig-C}&40.0&\cellcolor[HTML]{A9DFBF}41.4&\cellcolor[HTML]{A9DFBF}41.3&\cellcolor[HTML]{A9DFBF}42.9&\cellcolor[HTML]{A9DFBF}41.1&\multicolumn{2}{l}{\cellcolor[HTML]{A9DFBF} 41.8 (+1.8)}&\cellcolor[HTML]{A9DFBF}41.8&\cellcolor[HTML]{A9DFBF}42.6&\cellcolor[HTML]{A9DFBF}42.0&\cellcolor[HTML]{A9DFBF}41.5&\multicolumn{2}{l}{\cellcolor[HTML]{A9DFBF} 42.0 (+2.0)}\\
&\textit{Arg-I}&33.5&\cellcolor[HTML]{A9DFBF}37.1&\cellcolor[HTML]{A9DFBF}38.1&\cellcolor[HTML]{A9DFBF}37.6&\cellcolor[HTML]{A9DFBF}37.1&\multicolumn{2}{l}{\cellcolor[HTML]{A9DFBF} 37.6 (+4.1)}&\cellcolor[HTML]{A9DFBF}37.7&\cellcolor[HTML]{A9DFBF}37.9&\cellcolor[HTML]{A9DFBF}37.6&\cellcolor[HTML]{A9DFBF}36.9&\multicolumn{2}{l}{\cellcolor[HTML]{A9DFBF} 37.8 (+4.3)}\\
&\textit{Arg-C}&30.8&\cellcolor[HTML]{A9DFBF}34.3&\cellcolor[HTML]{A9DFBF}35.4&\cellcolor[HTML]{A9DFBF}34.7&\cellcolor[HTML]{A9DFBF}34.9&\multicolumn{2}{l}{\cellcolor[HTML]{A9DFBF} 34.8 (+4.0)}&\cellcolor[HTML]{A9DFBF}34.6&\cellcolor[HTML]{A9DFBF}34.5&\cellcolor[HTML]{A9DFBF}34.5&\cellcolor[HTML]{A9DFBF}34.1&\multicolumn{2}{l}{\cellcolor[HTML]{A9DFBF} 35.2 (+4.4)}\\
\cmidrule(lr){2-15}
&\textit{AVG}&40.7&\cellcolor[HTML]{A9DFBF}44.2&\cellcolor[HTML]{A9DFBF}44.5&\cellcolor[HTML]{A9DFBF}44.1&\cellcolor[HTML]{A9DFBF}43.5&\multicolumn{2}{l}{\cellcolor[HTML]{A9DFBF} 43.7 (+3.0)}&\cellcolor[HTML]{A9DFBF}44.5&\cellcolor[HTML]{A9DFBF}44.4&\cellcolor[HTML]{A9DFBF}44.3&\cellcolor[HTML]{A9DFBF}43.6&\multicolumn{2}{l}{\cellcolor[HTML]{A9DFBF} 44.5 (+3.8)}\\
\midrule
\multirow{7}{*}{\textit{Chinese}}
&\textit{Entity}&59.1&\cellcolor[HTML]{7DCEA0}67.8&\cellcolor[HTML]{52BE80}70.7&\cellcolor[HTML]{52BE80}70.7&\cellcolor[HTML]{52BE80}73.5&\multicolumn{2}{l}{\cellcolor[HTML]{52BE80} 73.5 (+14.4)}&\cellcolor[HTML]{7DCEA0}67.1&\cellcolor[HTML]{7DCEA0}68.8&\cellcolor[HTML]{52BE80}70.6&\cellcolor[HTML]{52BE80}70.2&\multicolumn{2}{l}{\cellcolor[HTML]{52BE80} 71.0 (+11.9)}\\
&\textit{Relation}&20.4&\cellcolor[HTML]{52BE80}31.2&\cellcolor[HTML]{52BE80}34.7&\cellcolor[HTML]{52BE80}35.9&\cellcolor[HTML]{52BE80}37.3&\multicolumn{2}{l}{\cellcolor[HTML]{52BE80} 37.8 (+17.4)}&\cellcolor[HTML]{52BE80}30.7&\cellcolor[HTML]{7DCEA0}28.2&\cellcolor[HTML]{7DCEA0}30.1&\cellcolor[HTML]{52BE80}35.6&\multicolumn{2}{l}{\cellcolor[HTML]{7DCEA0} 28.4 (+8.0)}\\
&\textit{Trig-I}&25.0&\cellcolor[HTML]{52BE80}48.6&\cellcolor[HTML]{52BE80}55.3&\cellcolor[HTML]{52BE80}56.2&\cellcolor[HTML]{52BE80}49.3&\multicolumn{2}{l}{\cellcolor[HTML]{52BE80} 52.5 (+27.5)}&\cellcolor[HTML]{52BE80}43.7&\cellcolor[HTML]{52BE80}53.5&\cellcolor[HTML]{52BE80}50.0&\cellcolor[HTML]{52BE80}50.7&\multicolumn{2}{l}{\cellcolor[HTML]{52BE80} 52.6 (+27.6)}\\
&\textit{Trig-C}&23.9&\cellcolor[HTML]{52BE80}45.6&\cellcolor[HTML]{52BE80}52.1&\cellcolor[HTML]{52BE80}52.0&\cellcolor[HTML]{52BE80}46.1&\multicolumn{2}{l}{\cellcolor[HTML]{52BE80} 49.0 (+25.1)}&\cellcolor[HTML]{52BE80}40.8&\cellcolor[HTML]{52BE80}50.0&\cellcolor[HTML]{52BE80}46.6&\cellcolor[HTML]{52BE80}47.4&\multicolumn{2}{l}{\cellcolor[HTML]{52BE80} 49.3 (+25.4)}\\
&\textit{Arg-I}&28.6&\cellcolor[HTML]{52BE80}42.6&\cellcolor[HTML]{52BE80}42.8&\cellcolor[HTML]{52BE80}40.9&\cellcolor[HTML]{52BE80}43.6&\multicolumn{2}{l}{\cellcolor[HTML]{52BE80} 42.3 (+13.7)}&\cellcolor[HTML]{52BE80}38.7&\cellcolor[HTML]{52BE80}39.6&\cellcolor[HTML]{52BE80}39.4&\cellcolor[HTML]{52BE80}39.8&\multicolumn{2}{l}{\cellcolor[HTML]{52BE80} 40.1 (+11.5)}\\
&\textit{Arg-C}&28.1&\cellcolor[HTML]{52BE80}40.3&\cellcolor[HTML]{52BE80}41.2&\cellcolor[HTML]{52BE80}39.4&\cellcolor[HTML]{52BE80}42.1&\multicolumn{2}{l}{\cellcolor[HTML]{52BE80} 40.8 (+12.7)}&\cellcolor[HTML]{7DCEA0}37.3&\cellcolor[HTML]{52BE80}38.4&\cellcolor[HTML]{52BE80}38.2&\cellcolor[HTML]{52BE80}38.2&\multicolumn{2}{l}{\cellcolor[HTML]{52BE80} 38.2 (+10.1)}\\
\cmidrule(lr){2-15}
&\textit{AVG}&30.9&\cellcolor[HTML]{52BE80}46.0&\cellcolor[HTML]{52BE80}49.5&\cellcolor[HTML]{52BE80}49.2&\cellcolor[HTML]{52BE80}48.7&\multicolumn{2}{l}{\cellcolor[HTML]{52BE80} 49.3 (+18.4)}&\cellcolor[HTML]{52BE80}43.1&\cellcolor[HTML]{52BE80}46.4&\cellcolor[HTML]{52BE80}45.8&\cellcolor[HTML]{52BE80}47.0&\multicolumn{2}{l}{\cellcolor[HTML]{52BE80} 46.6 (+15.7)}\\
\bottomrule
\end{tabularx}}
\vspace{-8pt}
\caption{Cross-lingual transfer experiments from English to target languages on three tasks: (1) NER on WikiAnn, (2) QA on TyDiQA-GoldP, and (3) Event extraction on ACE. Overall,  EasyProject (EProj.) achieves stronger performance compared to the alignment-based methods, and also outperforms using XML tags. ``Fine-tune$_{\text{en}}$'' refer to the zero-shot baselines, where the models are trained only on  English data. ``Ref'' column is the performance from prior work: QA   in \citet{hu2020xtreme} and NER from \citet{he2021effectiveness}; 
 ``XLM$_R$'' is our reimplementation using XLM-RoBERTa$_{\text{large}}$, which $\Delta$ is calculated against.  We show the results that use Google Translation (GMT) and NLLB model separately. For `-', the language is not supported by the supervised word aligner. Cells are colored by $\Delta$: \colorbox{red4}{\,\,\,\,\,\scriptsize{-10}}\colorbox{red3}{\,\,\,\,\,\,\,\,\scriptsize{-5}}\colorbox{red2}{\,\,\,\,\,\scriptsize{-1}}\colorbox{red1}{\,\,\,\scriptsize{0}}\colorbox{green1}{\,\,\,\textcolor{green1}{\scriptsize{0}}}\colorbox{green2}{\scriptsize{+1}\,\,\,\,\,}\colorbox{green3}{\scriptsize{+5}\,\,\,\,\,\,\,\,\,\,\,}\colorbox{green4}{\scriptsize{+10}\,\,\,\,\,\,\,\,\,\,\,}.}
\label{table:main_results}
\end{table*}

\begin{CJK*}{UTF8}{gbsn} 
\renewcommand{\arraystretch}{1.1}
\begin{table}[t!]
    \small
    \centering
    \resizebox{\linewidth}{!}{
    \begin{tabular}{@{\hspace{0.025cm}}r@{\hspace{0.025cm}}l@{\hspace{0.025cm}}}
    \toprule
       \textbf{English:}   &  He was buried in Woodlawn Cemetery in \underline{\smash{Bronx ,}}  \\
       &  \underline{\smash{New York City}} . \\
       \midrule
       \textbf{Alignment-based:}  & 他 被 埋葬 在 \underline{{\setlength{\fboxsep}{1pt}\colorbox{Blue!25}{纽约市\text{ }布朗}}\text{ }\textcolor{red}{克斯}}  的 伍德劳 恩公 墓 。  \\
        \midrule
       \textbf{EasyProject:}   & 他 被 埋葬 在 \underline{{\setlength{\fboxsep}{1pt}\colorbox{Blue!25}{纽约市\text{ }布朗\text{ }克斯}}}  的 伍德劳 恩公 墓 。 \\

       \bottomrule
       
    \end{tabular}
    }
    \vspace{-6pt}
    \caption{In this example, the correct projection should be ``\underline{纽约市\text{ }布朗\text{ }克斯}''. The outputs from two label projection methods are  {\setlength{\fboxsep}{1pt}\colorbox{Blue!25}{marked}}. For the alignment-based projection, ``克斯'' is \textcolor{red}{incorrectly missed}. The translations are based on Google Translation.}
    \label{tab:error_examples}
    \vspace{-15pt}
\end{table}
 \end{CJK*}

\vspace{-2.5pt}
\paragraph{Transfer Learning Results.} As summarized in Figure \ref{fig:bleu_f1}, EasyProject outperforms  alignment-based projection for most languages, even though span markers degrade translation quality. In Table \ref{table:masakha} and \ref{table:main_results}, we show that EasyProject almost always outperforms alignment-based projection on NER, QA, and the more challenging event extraction tasks, when training on a combination of English data and the translated  projected data in target languages. In addition, we  find that EasyProject  generally performs better than using XML tags, as the former has less impact on the translation quality.
We also notice the relatively low zero-shot performance in  \textit{ja}, \textit{zh}, and \textit{th} on WikiAnn dataset, which is consistent with scores reported in prior literature \cite{he2021effectiveness}. We suspect this is due to their distinct script systems, and adding EasyProject data brings significant improvements to all of them – Japanese (+25.5 F$_1$), Chinese (+18.8 F$_1$), and Thai (+14.4 F$_1$). 
 EasyProject (GMT) also improves the performance of  mT5$_{\text{large}}$ and mT5$_{\text{XL}}$ by 7.3 and 5.7 F$_1$ on average across all target languages, and mT5$_{\text{XXL}}$ by 2.2 F$_1$ on a subset of 8 languages, as shown in  Table \ref{table:mt5}. Full results of the mT5 model are provided in  Table \ref{table:mt5_detail} in  Appendix.
\vspace{-2.5pt}
\paragraph{Accuracy of Projected Annotations.} To answer why EasyProject can outperform  alignment-based method even though it degrades translation quality, we manually inspect  400 sentences sampled from the WikiANN training set. EasyProject correctly projects 100\% and 97.5\% of the label spans,  when using Google Translation and NLLB, respectively. Whereas the traditional method based on Awesome-align only achieves 97.5\% and 93.4\% accuracy.  We found EasyProject can  more accurately  preserve the boundaries of the label span. For the alignment-based method, most errors are caused by partial or missed alignments, as demonstrated in Table \ref{tab:error_examples}. More analyses are provided in the Appendix \ref{subsec:projection-rate-appendix}.

\vspace{-1pt}
\subsection{Size of Pre-training Data vs. Improvement in Performance}

\label{sec:pre-training-data-size}
\begin{figure}[t!]
    \centering
    \includegraphics[width=0.8\linewidth]{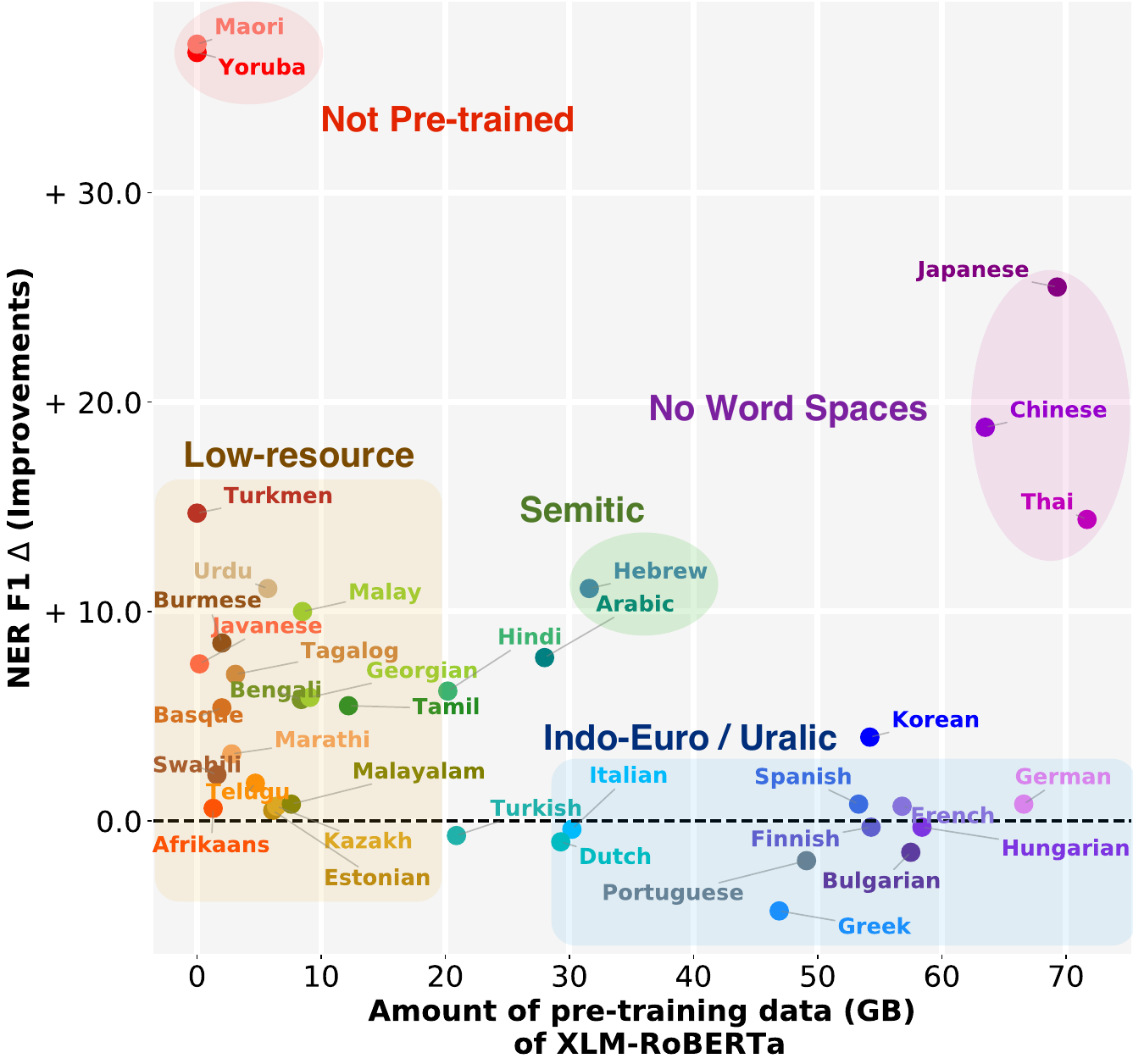}
    \vspace{-7pt}
    \caption{NER $\Delta$F$_1$  (EasyProject+GMT over  Fine-tune$_{\text{en}}$) vs. amount of pre-training data (GB) for XLM-RoBERTa$_{\text{large}}$.}
    \label{fig:ner_gb}
    \vspace{-18pt}
\end{figure}

Figure~\ref{fig:ner_gb} shows improvements in NER F$_1$  using EasyProject vs. size of data for each language 
in XLM-RoBERTa's pre-training corpus.
EasyProject provides larger improvements on low-resource languages and languages without whitespaces.
For high-resource languages in the Indo-European (e.g., Germanic and Romance) or Uralic families, using  projected data struggles to significantly improve over a strong fine-tuning baseline.

\vspace{-1pt}
\subsection{Transfer from non-English Languages}
\label{sec:multilingual}
\begin{figure*}[pht!]
    \centering
    \vspace{-5pt}
    \begin{subfigure}[t]{0.5\textwidth}
        \centering
                \includegraphics[width=0.8\textwidth]{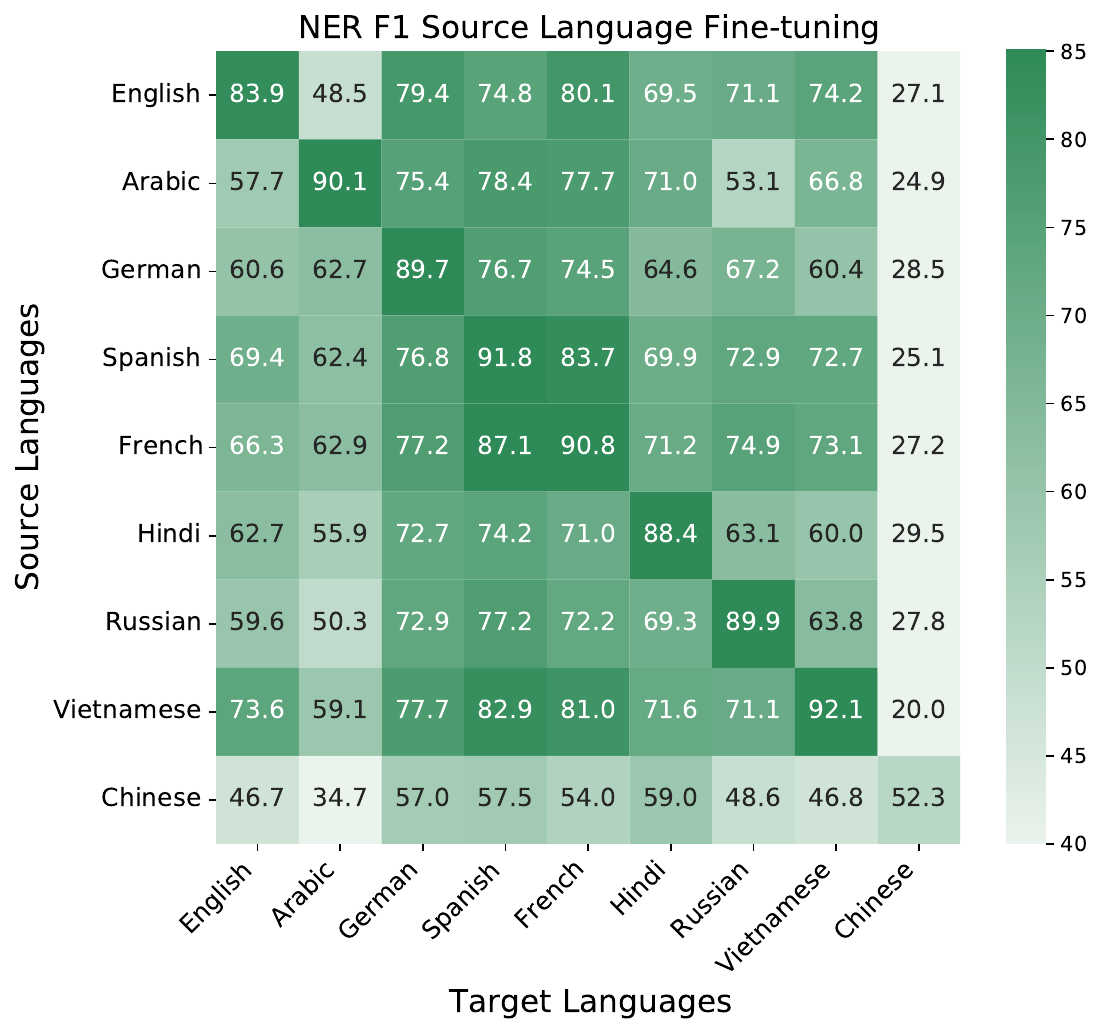}

    \end{subfigure}%
    ~ 
    \begin{subfigure}[t]{0.5\textwidth}
        \centering
        \includegraphics[width=0.8\textwidth]{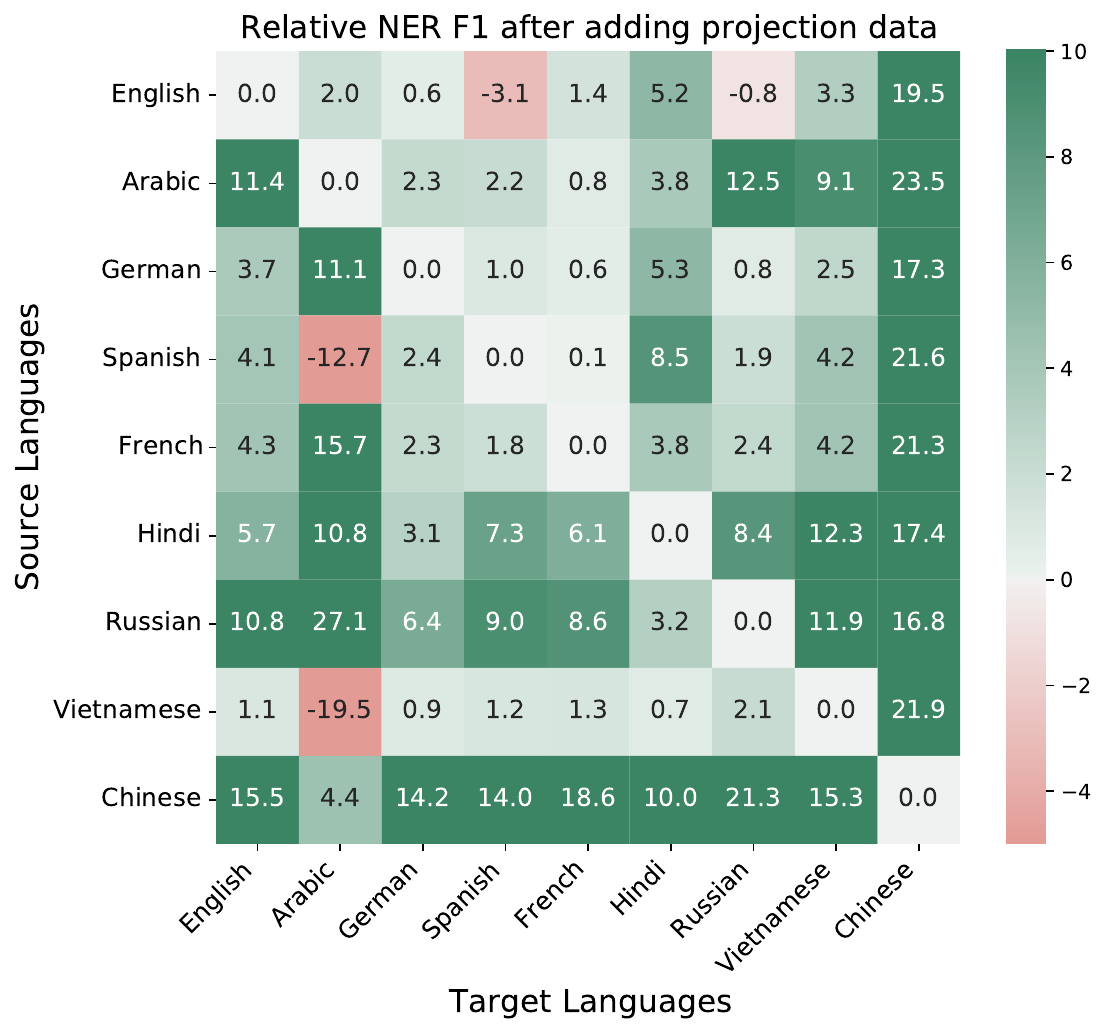}
    \end{subfigure}
    \vspace{-15pt}
    \caption{ (a) NER F$_1$ for fine-tuning on different source and target languages. (b) The relative difference of F$_1$ for models trained on \textit{source and projected data together} over \textit{source data only}, when using EasyProject with the  NLLB translation system. EasyProject  can be used for transfer learning from any language to any language provided with the availability of multilingual MT systems.}
    \vspace{-10pt}
    \label{fig:non_english}
\end{figure*}

Recent work has suggested that English may not always be the best language to transfer from \cite{turc2021revisiting}.
We demonstrate that the marker-based method is not limited to English-centric transfer learning; rather, it can be used for transfer learning from any language to any language provided with the availability of multilingual MT systems. In Figure~\ref{fig:non_english}, we show the relative F$_1$ improvements 
 of using EasyProject over fine-tuning  on source language only for 9 different languages (81 directions in total), leveraging the multilingual capabilities of NLLB. 
Fine-tuning models only on source-language data  does not work well when transferring to or from Chinese, consistent with observations from~\citet{hu2020xtreme}. 
The marker-based method addresses this problem by providing substantial improvements in F$_1$ on the WikiANN dataset for Chinese. Transferring to Arabic and from Russian are also challenging, but again, the marker-based method greatly boosts performance.

\vspace{-1pt}
\subsection{More Experiments and Analyses in the Appendix}
We also compare EasyProject against MulDA \citep{liu-2021-mulda} ($\S$\ref{sec:mulda}) and   bitext projection (\ref{sec:bitext}), as well as  evaluating it on low-resource languages: M\=aori and Turkmen ($\S$\ref{sec:low}). In addition, we analyze  the two branches of label projection methods from other aspects, including  projection rate  ($\S$\ref{subsec:projection-rate-appendix}) and translation speed  ($\S$\ref{appendix:translation_time}).  Due to space limits, we present all of them in the appendix.  On the data side, we  fix a sentence splitting issue for 12 extremely long sentences  in the ACE05 Arabic testset ($\S$\ref{appendix:fix_issue_ace} ). This issue is  noticed by other researchers \cite{huang2022multilingual} as well.
We will release the improved ACE Arabic  dataset to the community.

\vspace{-1pt}
\section{Conclusion}
In this paper, we presnet a thorough empirical assessment of various approaches for cross-lingual label projection. We also design an improved variant of the mark-and-translate method, which we call \textsc{EasyProject}. Experiments on  57 target languages and  three well-studied NLP tasks show that EasyProject consistently outperforms the alignment-based methods and effectively improves the performance of cross-lingual transfer.





\section*{Limitations}
While our study shows that EasyProject can effectively translate the source sentences with special markers inserted to the target languages, using the Google Translation and NLLB model, it is unclear whether all translation models can work well when special markers are inserted. 
To generalize this approach to future MT systems, we design a simple and computationally efficient approach to improve the robustness of MT systems in handling special markers.
However, the translation quality for the  marker-inserted text still falls behind the original text. We leave the work of further optimizing the translation quality as future work.

\section*{Acknowledgements}
This material is based upon work supported by the NSF (IIS-2052498) and IARPA via the BETTER and HIATUS programs (2019-19051600004, 2022-22072200004). The views and conclusions contained herein are those of the authors and should not be interpreted as necessarily representing the official policies, either expressed or implied, of NSF, ODNI, IARPA, or the U.S. Government. The U.S. Government is authorized to reproduce and distribute reprints for governmental purposes notwithstanding any copyright annotation therein.


\bibliography{anthology,custom}
\bibliographystyle{acl_natbib}

\newpage
\appendix

\clearpage



\section{Data statistics for CoNLL-2002/2003}
\label{sec:stat-for-conll0203}
The  statistics of the CoNLL-2002/2003  multilingual NER dataset are provided in Table \ref{table:stat_conll}.  


\renewcommand{\arraystretch}{1.2}
\begin{table}[h!]
\centering
\small
\vspace{-5pt}
\begin{tabular}{lc}
\toprule
& \textbf{CoNLL 2002/2003} \\
\midrule
\# of Lang. & 3 \\ 
\# of Docs & -- \\ 
\# of Sent. & 14k/3.2k/3.4k \\
Avg.  Length & --/14.5\\
Avg.  \# of Spans & 1.7 \\
\bottomrule
\end{tabular}
\caption{The detailed statistics of train/dev/test sets for CoNLL-2002/2003 dataset. \textbf{Avg. Length} represents the average number of tokens in each article/sentence, and \textbf{Avg.  \# of Spans} denotes the average number of annotated spans in each sentence.} 
\vspace{-15pt}
\label{table:stat_conll}
\end{table}

\section{Full Results on MasakhaNER2.0}
\label{sec:results-on-masakha2}

MasahkaNER2.0 is a NER dataset in the news domain, including the annotations on 20 African languages. Following the setting in the original paper \cite{adelani2022masakhaner}, we use CoNLL-03 dataset~\citep{sang2003conll} as the source corpus, and  train the mDeBERTv3~\citep{he2021debertav3} model on it. Then the trained model is evaluated on the test set of  MasahkaNER2.0, with a focus on the PER, ORG, and LOC types. 


\renewcommand{\arraystretch}{1.20}
\begin{table}[h!]
\small
\resizebox{\linewidth}{!}{
\begin{tabular}{lccccc}
\toprule
Language & Ref. & Fine-tune$_{\textit{en}}$ & +Awes. & +XML & +EasyProj. \\
\midrule
Bambara(bam)&38.4&37.1&45.0&44.3&45.8\\
Ghomala(bbj)&45.8&43.3&--&--&--\\
Ewe(ewe)&76.4&75.3&78.3&77.8&78.5\\
Fon(fon)&50.6&49.6&59.3&60.2&61.4\\
Hausa(hau)&72.4&71.7&72.7&71.6&72.2\\
Igbo(ibo)&61.4&59.3&63.5&59.6&65.6\\
Kinyarwanda(kin)&67.4&66.4&63.2&70.8&71.0\\
Luganda(lug)&76.5&75.3&77.7&77.9&76.7\\
Luo(luo)&53.4&35.8&46.5&50.0&50.2\\
Mossi(mos)&45.4&45.0&52.2&53.6&53.1\\
Chichewa(nya)&80.1&79.5&75.1&73.5&75.3\\
Naija(pcm)&75.5&75.2&--&--&--\\
chiShona(sna)&37.1&35.2&69.5&56.3&55.9\\
Kiswahili(swa)&87.9&87.7&82.4&81.7&83.6\\
Setswana(tsn)&65.8&64.8&73.8&72.9&74.0\\
Akan/Twi(twi)&49.5&50.1&62.7&64.7&65.3\\
Wolof(wol)&44.8&44.2&54.5&58.9&58.9\\
isiXhosa(xho)&24.5&24.0&61.7&71.9&71.1\\
Yoruba(yor)&40.4&36.0&38.1&36.8&36.8\\
isiZulu(zul)&44.7&43.9&68.9&74.8&73.0\\
\midrule
Averaged Perf. &56.9&55.0&63.2&63.8&64.3\\
\midrule
Averaged Proj. Rate & - & - & 86.9\% & 77.5\% &	93.7\% \\
\bottomrule
\end{tabular}
}
\caption{F1 scores on MasakhaNER2.0 using NLLB translation model. We skip  Ghomala and Naija as they are not supported by  NLLB.} 
\vspace{-15pt}
\label{table:results-on-masakha2}
\end{table}



\section{Details of Word Alignment Models}
\label{appendix:word_alignment}
\label{sec:implementation-details}
\paragraph{Awesome-align.} This aligner ~\citep{dou2021awesome}, when used in the unsupervised setting, primarily relies on the normalized similarity scores of all word pairs between the two sentences, which are calculated based on pre-trained multilingual word embeddings taken from specific Transformer layers. In the supervised setting, with access to parallel text, Awesome-align can be further improved by fine-tuning towards a set of self-training and language model objectives. We include experiments of both the unsupervised (\textit{Awesome}) and supervised (\textit{Awesome$_{ft}$}) versions of Awesome-align based on multilingual BERT, which has shown to achieve better word alignment results than XLM-RoBERTa$_{\text{base}}$.  
For the supervised version, we fine-tune an individual Awesome-align model for each of the 39 target languages in WikiANN using parallel sentences sampled from the M2M model's \cite{fan2021beyond} training datasets: CCAligned~\citep{el-kishky-etal-2020-ccaligned} and CCMatrix~\citep{schwenk-etal-2021-ccmatrix}.
Specifically, we randomly sample 200k parallel sentences from the CCAligned corpus for language pairs from English to \{\textit{te}, \textit{ka}, \textit{kk}, \textit{my}, \textit{th}, \textit{yo}\}, and the rest from the CCMatrix.

We use the codebase\footnote{\url{https://github.com/neulab/awesome-align}} from \citet{dou2021awesome} with the default softmax configuration to extract alignment. 
We do not apply the consistency optimization objective when fine-tuning the models because it may trade precision for recall, as suggested in the official instruction written by the authors.

\paragraph{QA-align.} This is a state-of-the-art  supervised approach~\citep{nagata2020supervised} that formulates the word alignment problem as a SQuAD-style question answering task by fine-tuning  multilingual BERT. Specifically, given a word in the source sentence, the model predicts the aligned span in the target sentence and reconciles both source-to-target and target-to-source directions by averaging and thresholding probabilities.  We trained the QA-align model for English to Arabic, German, French, Chinese, and Japanese, where gold annotated word alignment data is available.

We use the codebase from~\citet{nagata2020supervised}.\footnote{\url{https://github.com/nttcslab-nlp/word_align}} For the  training data of word alignment between  \textit{en} and \{\textit{de}, \textit{zh}, \textit{ja}, \textit{fr}\}, we use the same data as in \citet{nagata2020supervised}. For \textit{en}-\textit{ar}, we use the GALE English-Arabic word alignment data from LDC\footnote{LDC2014T05, LDC2014T10, LDC2014T14, LDC2014T19 LDC2013T10, LDC2013T14, LDC2014T03, LDC2014T08},  and use 80\% of the sentence pairs for training. The data statistics can be found in Table~\ref{table:qaalign}.



\renewcommand{\arraystretch}{1.20}
\begin{table}[t!]
\centering
\small
\begin{tabular}{lrrr}
\toprule
Lang. & Train & Test \\
\midrule
\textit{en-ar} &40,288&9,280\\
\textit{en-de} &300&208\\
\textit{en-fr} &300&147\\
\textit{en-ja} &653&357\\
\textit{en-zh} &4,879&610\\
\bottomrule
\end{tabular}
\vspace{-5pt}
\caption{Number of sentences in the  train/dev sets of the annotated  word alignment datasets. 
}
\vspace{-15pt}
\label{table:qaalign}
\end{table}

%
\section{Implementation Details of IE models}
\label{appendix:implementation_details_IE}
\label{appendix:implementation}
We follow the same learning rates and number of epochs reported in prior work: \citet{hu2020xtreme} for QA, \citet{he2021effectiveness} and \citet{pfeiffer2020madx} for NER (the latter for \textit{mi} and \textit{tk}) and  \citet{wu-2021-everything} for ACE.
For WikiANN NER~\citep{pan2017wikiann}, CoNLL-2002/2003 NER~\citep{sang2002conll,sang2003conll}, MasakhaNER2.0~\citep{adelani2022masakhaner}, and TyDiQA-GoldP~\citep{clark-etal-2020-tydi}, we use the codebase from the XTREME benchmark ~\citep{hu2020xtreme},\footnote{\url{https://github.com/google-research/xtreme}} and MasakhaNER2.0~\footnote{\url{https://github.com/masakhane-io/masakhane-ner}}, which is based on the Huggingface transformers  library~\citep{wolf2019huggingface}.
The hyperparameters of mDeBERTaV3 (276M) for MasakhaNER2.0 and XLM-RoBERTa$_{\text{large}}$(550M) and  for other datasets are presented in Table~\ref{table:hyperparameters} following~\citep{hu2020xtreme,he2021effectiveness,liu-2021-mulda,adelani2022masakhaner}.
We report the average result of three random seeds and select models based on the English development set. 

\renewcommand{\arraystretch}{1.2}
\begin{table}[t!]
\centering
\small
\resizebox{\linewidth}{!}{
\begin{tabular}{@{\hspace{0.15cm}}l@{\hspace{0.15cm}}c@{\hspace{0.15cm}}c@{\hspace{0.15cm}}c@{\hspace{0.15cm}}c@{\hspace{0.15cm}}}
\toprule

 & WikiANN & CoNLL &Masakha & TyDiQA-GoldP\\
\midrule
Task & NER & NER & NER & QA\\
Epochs & 5 & 10 & 5 & 3\\
Batch size & 32 & 32 & 32 & 8\\
Learning rate & 2e-5 & 2e-5 & 5e-5 & 3e-5\\
Warmup steps & 0 & 0 & 0 & 500\\
Weight decay & 0 & 0 & 0 & 0.0001\\
\bottomrule
\end{tabular}
}
\vspace{-5pt}
\caption{Hyperparameters for fine-tuning the NER and QA models.}
\label{table:hyperparameters}
\vspace{-18pt}
\end{table}

\paragraph{mT5 NER Model.}
Training  mT5$_{\text{XXL}}$~\citep{xue2021mt5} models, which have over 11 billion parameters, for the NER  task is computationally challenging. 
We formulate the WikiANN NER task as generating a sequence of tokens with special entity tags (e.g. \texttt{<per>, </per>}) inserted around the entity spans.
To fit the model into GPU memory for training, we freeze the embedding layer and the bottom 12 layers of both encoder and decoder during fine-tuning.
We also use the {\tt DeepSpeed}~\citep{deepspeed} ZeRo3 with 32-bits configurations. 
We first fine-tune the model on English data for 20 epochs with a learning rate of 1e-4.
To speed up the training process, we initialize the model from the English fine-tuned checkpoint and further fine-tune it on the combination of English and EasyProject + GMT data with a learning rate of 5e-5 for 5 epochs.
We report results of mT5$_{\text{large}}$ by averaging over three random seeds. We use one random seed for the XL and XXL models due to the heavy computing cost.
Experiment results of average performance across languages are shown in Table~\ref{table:mt5}, and results of each language are reported in Table~\ref{table:mt5_detail} in Appendix. 
\paragraph{ACE05.}
For ACE05 event extraction~\citep{walker2006ace}, 
we use the OneIE joint model v0.4.8 codebase\footnote{\url{http://blender.cs.illinois.edu/software/oneie/}} with the same hyperparameters as~\citet{wu-2021-everything}. 
For evaluation, we use the OneIE scoring tool to report F$_1$ scores for entities, relations, event triggers identification (Trig-I) and classification (Trig-C), argument role identification (Arg-I) and classification (Arg-C). 
We train models on the combination of English and  projected Chinese data from scratch in the Chinese experiment and select the model based on the English development set.
In the Arabic experiment, we initialize the model from the English fine-tuned checkpoint. We fine-tune the argument role classifier for event extraction tasks (Entity, Trig-I, Trig-C, Arg-I, Arg-C) and relation classifier in relation task for 5 epochs.  We set the learning rate of task-specific classifiers at 1e-6 and the encoder at 5e-4.
During the decoding process of relation classification, we only consider the joint model's relation and entity prediction scores.

On the Arabic data annotation side, the ACE Arabic data contains language-specific annotations on pronoun entities due to morphological stemming~\citep{zitouni2005arabic}, where we observe individual Arabic letter (prefix or suffix) is annotated as a pronoun entity.
Because such annotations don't exist in English data, the label projection process may cause inconsistency in translated-Arabic data.
Thus, we remove the pronoun entities in both Arabic test data and English training data for the Arabic experiment. The complete statistics of the Arabic test set is in Table~\ref{table:ace_arabic_test}. 
We report the average results of three random seeds.

\renewcommand{\arraystretch}{1.2}
\begin{table}[t!]
\centering
\small
\begin{tabular}{lrrrrr}
\toprule
 & \textit{en} & \textit{en}$^{\dagger}$ & \textit{zh} & \textit{ar} & \textit{ar}$^{\dagger}$\\
\midrule
Sent. & 19,216& 19,216 & 547 & 321 & 321\\
Entity & 47,554& 28,996 & 2,388 & 2,897 & 1,971\\
Relation & 7,159& 4,925 & 672 & 469 & 411\\
Trig. & 4,419& 3,125 & 190 & 232 & 232\\
Arg. & 6,607& 5,128 & 332 & 447 & 348\\
Tok/Sent & 14.2& 14.2 & 37.4 & 32.4 & 32.4\\
\bottomrule
\end{tabular}
\vspace{-5pt}
\caption{Statistic of the ACE05 English(\textit{en}) training set and Chinese(\textit{zh})/Arabic(\textit{ar}) test set. We hire an native Arabic  speaker to fix the Arabic test set by sentence-splitting the 12  articles that miss punctuation. 
$\dagger$: remove pronoun entities and related annotations in the events and relations.}
\vspace{-18pt}
\label{table:ace_arabic_test}
\end{table}

\section{Fixing Issues in the Arabic ACE Data}
\label{appendix:fix_issue_ace}
The  ACE data are pre-processed using the code from~\citet{lin2020oneie}. 
We use the same document splits as ~\citet{lin2020oneie} for English (ACE05-E$^{+}$) and Chinese (ACE05-CN). 
For Arabic, we use the document splits from~\citet{lan2020focused} following~\citet{wu-2021-everything}. 

In the processed Arabic test set from~\citet{wu-2021-everything}, we observed 12 extremely long sentences with an average length of 381 tokens, which are significantly longer than the rest of the sentences with an average length of 28.  This issue was also independently noticed by \citet{huang2022multilingual}.
A closer look reveals that these 12 sentences are 12 full articles in the original LDC release, which appear to be missing punctuation.
We hire a native Arabic  speaker to manually split them into sentences, resulting in 106 additional sentences. The data statistics are shown  in Table~\ref{table:ace_arabic_test}.
Because the ACE data is licensed, we will release the processing script instead.

\section{More Experiments}
In this section, we present more experiments to compare  EasyProject with other approaches.

\subsection{Comparison to MulDA}
\label{sec:mulda}
Table~\ref{table:MULDA} shows a direct comparison of EasyProject with MulDA~\citep{liu-2021-mulda}, another translation-based label projection approach that has been recently proposed for NER. MulDA replaces named entities with placeholders one by one, such as `\texttt{PER0}' and `\texttt{LOC1}', then invokes the MT system separately for each entity to translate and project the data. 
Thus, MulDA is  more time-consuming and costly than the EasyProject, which only requires one invocation of the MT system per sentence.
We find that EasyProject outperforms MulDA in German, Spanish, and Dutch at much less time cost. In this experiment, we follow MulDA's experimental setup, which uses the CoNLL NER dataset and trains only on the projected data.

In terms of translation speed, we calculate the relative time cost of EasyProject compared to translating the original sentences in CoNLL English data using the NLLB model on one A40 GPU. 
In Table~\ref{table:MULDA}, we observe marker-based (XML and \texttt{[]}) translation takes 1.2$\times$ and 1.3$\times$ longer of time to translate, due to the additional markers in both the input and output.
More analysis of the translation speed  is provided  in Appendix~\ref{appendix:translation_time}.

\renewcommand{\arraystretch}{1.15}
\begin{table}[ht!]
\centering
\small
\vspace{-5pt}
\begin{tabular}{llrrrrr}
\toprule
Method & MT & \textit{de} & \textit{es} & \textit{nl} & \textit{time}\\
\midrule
MulDA & GMT & 73.9 & 75.5 & 79.6 & -\\
MulDA & NLLB & 74.5 & 73.5 & 77.5 &  2.4$\times$\\
\midrule
XML & GMT & 74.3 & \textbf{77.1} & 79.8 &-\\

XML & NLLB & \textbf{75.3} & 74.9 &78.3&1.2$\times$\\
\midrule
EasyProject & GMT & 74.9 &70.3 &\textbf{79.9}&-\\
EasyProject & NLLB & 75.2 &73.0 &77.5 &1.3$\times$\\

\bottomrule
\end{tabular}
\vspace{-5pt}
\caption{Comparison of MulDA~\citep{liu-2021-mulda} and EasyProject on CoNLL NER (F$_1$), using \textit{projected data only}. ``\textit{time}'': relative time cost compared to translating the original sentence.}
\label{table:MULDA}
\vspace{-10pt}
\end{table}

\subsection{Comparison to Bitext Projection}
\label{sec:bitext}


Besides translation-projection, another alternative is bitext-projection, in which bilingual parallel corpora are used in place of a machine translation system. For example, we can apply a trained English IE model to the English side of the bilingual parallel corpus, then use word alignment to project the automatically predicted labels to the corresponding sentences in  target languages.

In Table~\ref{table:bitext}, we show that bitext-projection improves F$_1$ of WikiANN NER on 6 out of 8  languages used in ~\cite{wu-2021-everything} over the fine-tuning baseline (Fint-tune$_{\textit{en}}$), but is outperformed by EasyProject. For this experiment, we randomly sample 100,000 parallel sentences for each of the eight languages  from WikiMatrix~\citep{schwenk2019wikimatrix}, an automatically mined bitext corpus from Wikipedia that matches the domain of WikiANN. We use an XLM-RoBERTa$_{\text{large}}$ NER model trained on WikiANN English data  with 83.9 F$_1$ to generate named entity labels, and then apply Awesome-align to project labels to the target language. Finally, we train the XLM-RoBERTa$_{\text{large}}$ model on English and bitext-projected data together for 2 epochs (Bitext$_{100k}$).


Bitext$_{100k}$ loses 13.9 F$_1$ score for Vietnamese (\textit{vi}), most likely due to Awesome-align projection errors being magnified by fine-tuning on 100,000 projected sentences.
One surprising finding is that the Bitext$_{100k}$ improves by an absolute 4.4 F$_1$ score on Spanish and 1.1 F$_1$ on Russian.  Translation-projection approaches struggle on these two languages as shown in Table~\ref{table:main_results}.

\renewcommand{\arraystretch}{1.15}
\begin{table}[ht]
\centering
\small
\vspace{-5pt}
\resizebox{\linewidth}{!}{
\begin{tabularx}{0.5\textwidth} { 
   >{\hsize=1.2\hsize}X
   >{\hsize=1.5\hsize}X
   >{\hsize=4.9\hsize}X
   >{\hsize=4.8\hsize}X
   >{\hsize=4.9\hsize}X }
\toprule
\multirow{2}{*}{Lang.} & \multirow{2}{*}{FT$_{\textit{en}}$} & \multirow{2}{*}{EasyProject} & \multirow{2}{*}{Bitext$_{\textit{100k}}$} & EasyProject\\
&&&&+Bitext$_{\textit{100k}}$\\
\midrule
\textit{ar} & 48.5 & \cellcolor[HTML]{7DCEA0} 56.3 (+7.6)& \cellcolor[HTML]{A9DFBF}52.6 (+4.1)& \cellcolor[HTML]{A9DFBF}51.3 (+2.8)\\
\textit{de} & 79.4 & \cellcolor[HTML]{D4EFDF }80.2 (+0.8)& \cellcolor[HTML]{A9DFBF}81.0 (+1.6)& \cellcolor[HTML]{A9DFBF}81.4 (+2.0)\\
\textit{es} & 74.8 & \cellcolor[HTML]{D4EFDF }75.6 (+0.8) & \cellcolor[HTML]{A9DFBF}79.2 (+4.4)& \cellcolor[HTML]{A9DFBF}77.7 (+2.9)\\
\textit{fr} & 80.1 & \cellcolor[HTML]{D4EFDF }80.8 (+0.7) & \cellcolor[HTML]{D4EFDF }80.5 (+0.4)& \cellcolor[HTML]{A9DFBF}82.4 (+2.3)\\
\textit{hi} & 69.5 & \cellcolor[HTML]{7DCEA0}75.7 (+6.2)& \cellcolor[HTML]{FDEDEC}68.6 (-0.9)& \cellcolor[HTML]{7DCEA0}74.6 (+5.1)\\
\textit{ru} & 71.1 & \cellcolor[HTML]{FADBD8}68.2 (-2.9)& \cellcolor[HTML]{D4EFDF }72.2 (+1.1)& \cellcolor[HTML]{FADBD8}68.6 (-2.5)\\
\textit{vi} & 74.2 & \cellcolor[HTML]{A9DFBF}76.0 (+1.8)& \cellcolor[HTML]{F5B7B1}60.3 (-13.9)& \cellcolor[HTML]{A9DFBF}77.5 (+3.3)\\
\textit{zh} & 27.1 & \cellcolor[HTML]{52BE80}45.9 (+18.8) & \cellcolor[HTML]{A9DFBF}31.7 (+4.6)& \cellcolor[HTML]{52BE80}44.5 (+17.4)\\
\midrule
AVG &65.6&\cellcolor[HTML]{A9DFBF}69.8 (+4.2)&\cellcolor[HTML]{D4EFDF }65.8 (+0.2)& \cellcolor[HTML]{A9DFBF} 69.8 (+4.2)\\
\bottomrule

\end{tabularx}
}
\vspace{-5pt}
\caption{Comparison of NER F$_1$ on WikiANN between Bitext-Projection with 100,000 bilingual sentence pairs and  EasyProject with GMT.}
\label{table:bitext}
\vspace{-10pt}
\end{table}

\renewcommand{\arraystretch}{1.2}
\begin{table}[pt!]
\centering
\small
\setlength{\tabcolsep}{1pt}
\begin{tabular}{lccc}
\toprule
Method & MT & M\=aori(\textit{mi}) & Turkmen(\textit{tk})\\
\midrule
mBERT$^{\dagger}$ & - & 21.8 & 47.2\\
XLM-RoBERTa$_{\text{base}}^{\dagger}$ & - &15.9 & 43.4\\
\midrule
XLM-RoBERTa$_{\text{large}}$  & - & 30.3 &52.2 \\
~+~word-translation & PanLex&42.5&53.8\\
~+~Awesome-align & GMT & 46.1 & \textbf{60.7} \\
~+~EasyProject & GMT & \textbf{53.0} & 58.1\\
\bottomrule
\end{tabular}
\vspace{-5pt}
\caption{F$_1$ scores for cross-lingual NER from English to two very \textbf{low-resource languages} on WikiANN. PanLex is a bilingual dictionary. $\dagger$: English fine-tuning results reported in \citep{pfeiffer2020madx}.}
\label{table:low}
\vspace{-15pt}
\end{table}

\subsection{Experiments on Low-resource Languages}
\label{sec:low}

\renewcommand{\arraystretch}{1.2}
\begin{table*}[t!]
\centering
\small
\resizebox{\linewidth}{!}{
\begin{tabular}{ccccccc|ccccccc}
\toprule
\multirow{2}{*}{Language}  & \multicolumn{3}{c}{NLLB} & \multicolumn{3}{c|}{NLLB$_{\text{finetune}}$} & \multirow{2}{*}{Language} & \multicolumn{3}{c}{NLLB} & \multicolumn{3}{c}{NLLB$_{\text{finetune}}$} \\
\cmidrule(lr){2-4} \cmidrule(lr){5-7} \cmidrule(lr){9-11} \cmidrule(lr){12-14}
& \text{Orig} & \text{XML} & {\tt []} & \text{Orig} & \text{XML} & {\tt []} & & \text{Orig} & \text{XML} & {\tt []} & \text{Orig} & \text{XML} & {\tt []}\\
\midrule
Afrikaans(af)&44.0&44.3&43.6&45.9&45.4&45.5&Luo(luo)&15.6&15.6&15.3&16.5&16.0&15.9\\
Arabic(ar)&39.8&38.5&38.2&39.0&36.7&37.6&Malayalam(ml)&34.1&33.2&33.4&39.1&36.6&37.8\\
Bulgarian(bg)&47.8&47.5&46.7&48.4&45.0&46.7&Mossi(mos)&6.4&6.0&6.4&6.5&6.3&6.4\\
Bambara(bm)&10.5&10.5&10.3&10.4&10.0&10.2&Marathi(mr)&29.4&27.9&27.8&29.8&27.1&28.1\\
Bengali(bn)&34.6&32.6&33.2&35.3&31.1&33.0&Malay(ms)&45.0&43.9&43.5&45.1&43.6&43.5\\
German(de)&45.3&44.2&44.1&45.7&43.3&44.6&Burmese(my)&16.2&16.1&14.0&20.4&16.0&17.7\\
Ewe(ee)&16.3&16.0&16.4&16.7&15.9&16.2&Dutch(nl)&35.2&34.9&34.4&35.0&33.0&33.8\\
Greek(el)&37.6&36.7&36.2&37.1&34.6&35.5&Chichewa(ny)&17.4&17.6&17.0&20.5&19.5&19.9\\
Spanish(es)&32.7&32.3&32.0&32.1&31.0&31.3&Portuguese(pt)&53.7&53.0&52.5&54.6&52.2&53.1\\
Estonian(et)&33.5&32.9&32.3&33.0&30.5&31.3&Russian(ru)&40.2&38.8&39.5&40.1&36.9&38.9\\
Basque(eu)&26.1&26.3&24.9&30.8&28.8&29.2&Kinyarwanda(rw)&24.6&24.2&22.7&26.9&24.6&25.7\\
Persian(fa)&32.4&31.8&31.4&32.3&30.3&30.8&Shona(sn)&19.4&18.2&18.2&19.6&16.6&17.5\\
Finnish(fi)&32.4&31.9&32.0&32.9&29.5&31.2&Swahili(sw)&37.6&37.1&37.4&40.1&38.2&39.1\\
Benin(fon)&5.3&5.0&5.7&5.3&4.0&5.6&Tamil(ta)&36.4&34.6&34.2&37.7&34.6&36.3\\
French(fr)&55.0&54.7&53.6&55.3&52.9&53.6&Telugu(te)&38.3&37.3&37.5&39.1&37.4&38.2\\
Hausa(ha)&29.2&28.6&28.0&29.4&28.0&28.5&Thai(th)&32.2&30.7&28.3&32.9&28.8&29.9\\
Hebrew(he)&41.2&39.6&38.9&39.4&35.6&37.1&Tagalog(tl)&36.9&35.1&35.4&34.8&32.8&33.4\\
Hindi(hi)&40.9&39.1&40.0&41.3&38.3&40.6&Tswana(tn)&25.8&25.8&25.2&26.5&24.5&26.5\\
Hungarian(hu)&35.5&34.7&34.1&35.5&33.0&33.8&Turkish(tr)&40.4&39.2&38.7&41.0&37.7&38.9\\
Indonesian(id)&46.8&45.7&45.7&46.5&43.9&45.3&Twi(tw)&16.1&16.1&15.8&16.7&16.5&16.2\\
Igbo(ig)&19.7&19.8&19.3&20.2&19.5&20.0&Urdu(ur)&30.8&29.5&29.8&30.6&29.0&29.8\\
Italian(it)&37.1&36.2&36.0&36.0&33.9&34.6&Vietnamese(vi)&42.1&41.5&41.3&41.9&39.8&40.8\\
Japanese(ja)&17.8&17.0&13.7&19.9&18.0&18.8&Wolof(wo)&9.2&9.2&9.4&9.1&9.7&9.2\\
Javanese(jv)&28.9&28.1&28.3&29.4&27.9&28.7&Xhosa(xh)&24.2&23.8&22.9&26.5&23.8&24.9\\
Georgian(ka)&32.2&31.2&31.0&32.6&28.7&31.3&Yoruba(yo)&9.0&10.4&8.5&6.8&6.4&8.0\\
Kazakh(kk)&30.6&30.1&29.9&33.2&30.1&31.4&Chinese(zh)&23.9&24.2&22.9&28.0&26.2&26.8\\
Korean(ko)&24.9&23.9&23.6&25.2&22.2&24.4&Zulu(zu)&30.3&29.1&29.2&30.8&27.8&28.5\\
Ganda(lg)&12.5&13.0&12.4&13.0&13.2&13.1&\\ \midrule
\textbf{Average}&30.2&29.5&29.1&30.9&28.8&29.7\\
\bottomrule
\end{tabular}
}
\vspace{-5pt}
\caption{BLEU score of NLLB on FLORES-200~\citep{nllb2022} dev set (1000 sentences per language). We compare three types of translations: original translation (Orig), inserted with XML and {\tt []} special markers. We also fine-tuned NLLB with different markers using the  method described in Appendix \ref{sec:fine-tune-nllb}. We found that the fine-tuned NLLB model using square brackets has the least negative impact on translation quality}
\label{table:bleu_corpus_full}
\vspace{-15pt}
\end{table*}

To investigate the effectiveness of label projection on very low-resource languages~\citep{pfeiffer2020madx}, we conduct experiments on M\=aori (\textit{mi}) and Turkmen (\textit{tk}), which are not covered by the pre-trained language models (i.e., XLM-RoBERTa and mBERT) and have a small number of Wikipedia articles ($\sim$1.2k for M\=aori and $\sim$0.5k for Turkman). 
As shown in Table~\ref{table:low}, EasyProject improves F$_1$ score by an absolute 22.7 F$_1$ score on M\=aori and 5.9 F$_1$ on Turkmen compared to fine-tuning  on English data only. 
We also include a lexicon-based baseline, replacing English words with their word-to-word translations based on PanLex \cite{kamholz-etal-2014-panlex}, a commonly used multilingual dictionary. 
Both EasyProject and Awesome-align significantly outperform the word-level translations, likely because word-level translations still follow the English word orders and fail to capture the variation of word orders in M\=aori and Turkmen. 
For example, M\=aori has a verb-subject-object word order, while Turkmen uses a subject-object-verb. 
The improvement is less significant on Turkmen than M\=aori, potentially because Turkmen is close to Turkish, which is covered by both mBERT and XLM-RoBERTa. 
This is also a plausible reason why Awesome-align that uses mBERT did better on Turkmen.

\section{More  Analysis on EasyProject}
\label{appendix:translation-evaluation}
Here, we present more analysis of the EasyProject method in comparison with the traditional  pipeline approach based on word alignment.
\subsection{Translation Quality}
\label{appendix:translation-quality-evaluation}



To further measure the impact of adding special markers on the translation quality for the NLLB model, we adopt the evaluation setup used by NLLB~\citep{nllb2022} which utilizes the professional human-translated FLORES-200 parallel corpora (1000 sentences per language).
For the  marker-based approaches (``XML'' and ``{\tt []}''),  special markers are removed from the outputs before calculating the BLEU scores.  Table~\ref{table:bleu_corpus_full} presents the BLEU scores for the original NLLB model (3.3B) and the  NLLB model further fine-tuned with three types of parallel sentences (original, inserted with XML and {\tt []} markers).

As there is no gold NER annotation on the parallel corpus, we first train a NER model based on XLM-RoBERT$_{\text{large}}$  on the WikiAnn dataset, achieving an F$_1$ score of 83.9. We then apply the trained model to the English side of the parallel corpus and apply the EasyProject method to translate sentences into the target language.
After removing the special markers from the translation outputs, we use the {\tt sacreBLEU}\footnote{\url{https://github.com/mjpost/sacrebleu}} to calculate the BLEU scores by comparing the translations against gold references. We follow the NLLB evaluation setting and use  multilingual tokenizations (flores200).\footnote{\url{https://github.com/mjpost/sacrebleu/blob/master/CHANGELOG.md}}

\subsection{Projection Rate} 
\label{subsec:projection-rate-appendix}

We then compare the projection rate for all label projection methods, for which we divide the number of annotations after projection by the number of annotations occurring in the original training data.  We also include the average number of successfully projected sentences after filtering out the incorrect ones, which have a different number of annotations compared to the source sentence. For example, the source sentence has a {\tt LOC} and a {\tt PER} entity, but the projected sentence has two {\tt LOC} entities. Such sentences will be filtered.
For QA-align in WikiANN NER, we show the average statistics for 5 languages \{\textit{ar, de, fr, ja, zh}\} that have supervised training data.

As shown in Table \ref{table:projection_rate}, Google Translation (GMT) is very robust in handling special markers, and EasyProject has a nearly perfect 100\% projection rate, higher than any word alignment-based method. Our manual inspection of 100 sentences, randomly sampled from the WikiANN training set for English to Chinese projection, also reveals that GMT+EasyProject successfully projects all the sentences without mistakes on any target named entities, whereas Awesome-align only projected 94 sentences and caused 4 entity projection errors. 
According to our manual analysis, EasyProject is less likely to introduce errors than the word alignment-based method because the use of special markers encourages full-span projection. 

We found that most errors are caused by partial or missed alignments, which often occur when a span contains multiple words, a sentence contains many spans, or when both a Chinese transliteration and the original English name occur together in the translated sentence, which is a correct way to translate but poses challenges for label projection. More examples of alignment errors can be found in Table~\ref{tab:error_examples_more} in Appendix.

\begin{CJK*}{UTF8}{gbsn} 
\renewcommand{\arraystretch}{1.1}
\begin{table}[t!]
    \small
    \centering
    \resizebox{\linewidth}{!}{
    \begin{tabular}{@{\hspace{0.025cm}}r@{\hspace{0.025cm}}l@{\hspace{0.025cm}}}
    \toprule

        \textbf{English} \textbf{\#1:}   & \underline{Dean of Wolverhampton} ( 1373 - 1394 )   \\
       \midrule
       \textbf{Alignment-based:}  &\underline{\textcolor{red}{沃尔弗}\text{ } {\setlength{\fboxsep}{1pt}\colorbox{Blue!25}{汉普顿\text{ }伯爵}}}, 1373 - 1394 年    \\
       \midrule
       \textbf{EasyProject:}  & \underline{{\setlength{\fboxsep}{1pt}\colorbox{Blue!25}{沃尔弗\text{ }汉普顿\text{ }伯爵}}} 1373 - 1394   \\
       \midrule
       \midrule
             \textbf{English} \textbf{\#2:}  & \underline{Pino Daniele} ( 1955 - 2015 )   \\
       \midrule
       
       \textbf{Alignment-based:}  &\underline{\textcolor{red}{皮诺 · 丹尼 埃尔}} ({\setlength{\fboxsep}{1pt}\colorbox{Blue!25}{Pino Daniele}} , 1955 - 2015 年 )   \\
       \midrule
       \textbf{EasyProject:}  & \underline{{\setlength{\fboxsep}{1pt}\colorbox{Blue!25}{皮诺 · 丹尼尔}}} (1955 - 2015 年 )   \\
       
       \bottomrule
       
    \end{tabular}
    }
    \vspace{-5pt}
    \caption{Examples from WikiANN dataset using NLLB translation. The {\setlength{\fboxsep}{1pt}\colorbox{Blue!25}{outputs}} from two projection methods and \underline{correct answers} are  marked. In \#1, the alignment-based method \textcolor{red}{incorrectly misses} the ``沃尔弗'', which is a part of the translation for ``Wolverhampton''. In \#2, for the alignment-based method,  the Chinese translation (``皮诺· 丹尼埃尔'') and the original English span (``Pino Daniele'')   occur together in the translation. Alignment-based method \textcolor{red}{incorrectly misses} the correct projection ``皮诺· 丹尼埃尔'' and project to ``{\setlength{\fboxsep}{1pt}\colorbox{Blue!25}{Pino Daniele}}''. } 
    \label{tab:error_examples_more}
    \vspace{-10pt}
\end{table}
\end{CJK*}

\begin{table}[t!]
\centering
\small

\begin{tabular}{lccc}
\toprule
 & \#Inputs  & \#Outputs & Time (sec)\\
\midrule
Original & 279,678 & 335,963 & 4,452\\
XML & 460,002 & 468,815 & 5,486\\
\texttt{[]} & 326,294 & 379,796 & 4,107\\
Entity & 64,293 & 72,309 & 1,553\\
\bottomrule
\end{tabular}
\vspace{-5pt}
\caption{Number of tokens in the three types of input sentences: original CoNLL NER English training data, adding XML and {\tt []} special markers; and their corresponding  translations  in German. Time is the total translation clockwise time in seconds.}
\label{table:marker_time}
\vspace{-20pt}
\end{table}

\subsection{Translation Speed}
\label{appendix:translation_time}

Additional special markers added to the source sentence will affect the translation speed.
In Table~\ref{table:marker_time}, we show the number of tokens in the input and translation output. We use  the CoNLL-2002/2003 NER English training set as the source sentences, and translate them into German. All sentences are tokenized by the NLLB tokenizer.
We also show the translation time per sentence and per entity  on an A40 GPU with a batch size of 32. 

We estimate using XML tags takes 1.2$\times$  time compared to translating the original sentences, and EasyProject takes 1.3$\times$ time as it requires the additional translation of each entity span, for identifying the label correspondence.

%

\begin{table*}[pht!]
\centering
\small
\renewcommand{\arraystretch}{1.1}
\begin{tabularx}{\textwidth}{>{\hsize=0.5\hsize}Xlc*{10}{Y}}
\toprule
& & \multicolumn{3}{c}{NLLB+Word Aligner} & \multicolumn{2}{c}{NLLB+Markers} & \multicolumn{3}{c}{GMT+Word Aligner} & \multicolumn{2}{c}{GMT+Markers}\\
\cmidrule(lr){3-5} \cmidrule(lr){6-7} \cmidrule(lr){8-10} \cmidrule(lr){11-12} 
& & {\scriptsize QAalign} & {\scriptsize Awesome.} & {\scriptsize Awesome$_{\text{ft}}$} & {\scriptsize XML}  & \scriptsize{EProj.}  & {\scriptsize QAalign} & {\scriptsize Awesome.} & {\scriptsize Awesome$_{\text{ft}}$} & {\scriptsize XML}  & \scriptsize{EProj.}\\
\midrule
\multirow{2}{*}{NER} 
&\# Sents & 18,486(5)& 18,274	&18,587 &13,959	&19,470 & 19,187(5)	&19,003	&19,408	&20,000	&20,000\\
&Proj. Rate & 92.4(5) & 91.4	&92.9&	69.8 & 97.4 & 96.0(5) & 94.8 & 98.2 & 100 & 100 \\
\midrule
\multirow{2}{*}{{Event}} 
&\# Sents & 15,491&	14,840	&15,857	&7,308	&16,846 &16,264	&16,631&	16,903&	19,185&	19,185\\
&Proj. Rate & 80.6&	77.2 & 82.5 & 38.0 & 87.7 & 90.4&	92.6&	93.6&	99.9&	99.9\\
\midrule
\multirow{2}{*}{QA} 
&\# Sents & - & 3,613	&3,654& 1,573	&3,564 & -	& 3,623	& 3,649	&3,695	&3,695			\\
&Proj. Rate & - & 97.8 & 98.9 & 42.6 & 96.4& - & 97.8 &	99.1 & 100 & 100 \\
\bottomrule
\end{tabularx}
\vspace{-5pt}
\caption{Diagnosis analysis of projected data based on two metrics: number of sentences and the percentage of the projected annotations (Proj. Rate). For QA-align in NER, we  show 5 languages \{{\it ar,de,fr,ja,zh}\}.}
\label{table:projection_rate}
\end{table*}

\renewcommand{\arraystretch}{1.2}
\begin{table*}[ht!]
\centering
\scriptsize
\begin{tabular}{lcccccccccccc}
\toprule
Lang. & XLM-R$_{\text{large}}$ & \multicolumn{2}{c}{+EasyProject} & mT5$_{\text{large}}$ & \multicolumn{2}{c}{+EasyProject} & mT5$_{\text{XL}}$ & \multicolumn{2}{c}{+EasyProject} & mT5$_{\text{XXL}}$ & \multicolumn{2}{c}{+EasyProject}\\
\midrule\textit{af}&78.6&\multicolumn{2}{l}{\cellcolor[HTML]{E9F7EF} 79.2 (+0.6)}&79.2&\multicolumn{2}{l}{\cellcolor[HTML]{A9DFBF} 81.0 (+1.8)}&77.2&\multicolumn{2}{l}{\cellcolor[HTML]{A9DFBF} 79.6 (+2.4)}&-&\multicolumn{2}{l}{-}\\
\textit{ar}&48.5&\multicolumn{2}{l}{\cellcolor[HTML]{7DCEA0} 56.3 (+7.8)}&53.1&\multicolumn{2}{l}{\cellcolor[HTML]{52BE80} 66.1 (+13.0)}&57.4&\multicolumn{2}{l}{\cellcolor[HTML]{52BE80} 68.0 (+10.6)}&62.2&\multicolumn{2}{l}{\cellcolor[HTML]{A9DFBF} 66.1 (+3.9)}\\
\textit{bg}&82.1&\multicolumn{2}{l}{\cellcolor[HTML]{FADBD8} 80.6 (-1.5)}&58.5&\multicolumn{2}{l}{\cellcolor[HTML]{52BE80} 77.0 (+18.5)}&61.5&\multicolumn{2}{l}{\cellcolor[HTML]{52BE80} 76.1 (+14.6)}&-&\multicolumn{2}{l}{-}\\
\textit{bn}&75.1&\multicolumn{2}{l}{\cellcolor[HTML]{7DCEA0} 80.9 (+5.8)}&57.3&\multicolumn{2}{l}{\cellcolor[HTML]{52BE80} 76.2 (+18.9)}&65.7&\multicolumn{2}{l}{\cellcolor[HTML]{52BE80} 75.8 (+10.1)}&-&\multicolumn{2}{l}{-}\\
\textit{de}&79.4&\multicolumn{2}{l}{\cellcolor[HTML]{E9F7EF} 80.2 (+0.8)}&75.6&\multicolumn{2}{l}{\cellcolor[HTML]{A9DFBF} 77.9 (+2.3)}&75.9&\multicolumn{2}{l}{\cellcolor[HTML]{A9DFBF} 77.6 (+1.7)}&76.5&\multicolumn{2}{l}{\cellcolor[HTML]{E9F7EF} 77.3 (+0.8)}\\
\textit{el}&79.3&\multicolumn{2}{l}{\cellcolor[HTML]{FADBD8} 75.0 (-4.3)}&61.6&\multicolumn{2}{l}{\cellcolor[HTML]{52BE80} 81.6 (+20.0)}&79.4&\multicolumn{2}{l}{\cellcolor[HTML]{FADBD8} 77.2 (-2.2)}&-&\multicolumn{2}{l}{-}\\
\textit{es}&74.8&\multicolumn{2}{l}{\cellcolor[HTML]{E9F7EF} 75.6 (+0.8)}&85.7&\multicolumn{2}{l}{\cellcolor[HTML]{A9DFBF} 87.0 (+1.2)}&86.3&\multicolumn{2}{l}{\cellcolor[HTML]{FADBD8} 85.3 (-1.0)}&85.6&\multicolumn{2}{l}{\cellcolor[HTML]{E9F7EF} 86.4 (+0.8)}\\
\textit{et}&79.6&\multicolumn{2}{l}{\cellcolor[HTML]{E9F7EF} 80.1 (+0.5)}&71.8&\multicolumn{2}{l}{\cellcolor[HTML]{A9DFBF} 72.8 (+1.0)}&71.7&\multicolumn{2}{l}{\cellcolor[HTML]{A9DFBF} 73.2 (+1.4)}&-&\multicolumn{2}{l}{-}\\
\textit{eu}&63.6&\multicolumn{2}{l}{\cellcolor[HTML]{7DCEA0} 69.0 (+5.4)}&64.0&\multicolumn{2}{l}{\cellcolor[HTML]{A9DFBF} 68.0 (+4.0)}&64.0&\multicolumn{2}{l}{\cellcolor[HTML]{52BE80} 74.1 (+10.1)}&-&\multicolumn{2}{l}{-}\\
\textit{fa}&64.3&\multicolumn{2}{l}{\cellcolor[HTML]{F1948A} 52.0 (-12.3)}&47.0&\multicolumn{2}{l}{\cellcolor[HTML]{52BE80} 67.5 (+20.5)}&46.1&\multicolumn{2}{l}{\cellcolor[HTML]{52BE80} 64.9 (+18.8)}&-&\multicolumn{2}{l}{-}\\
\textit{fi}&80.6&\multicolumn{2}{l}{\cellcolor[HTML]{FDEDEC} 80.3 (-0.3)}&74.6&\multicolumn{2}{l}{\cellcolor[HTML]{A9DFBF} 78.0 (+3.4)}&73.5&\multicolumn{2}{l}{\cellcolor[HTML]{7DCEA0} 79.2 (+5.7)}&-&\multicolumn{2}{l}{-}\\
\textit{fr}&80.1&\multicolumn{2}{l}{\cellcolor[HTML]{E9F7EF} 80.8 (+0.7)}&84.6&\multicolumn{2}{l}{\cellcolor[HTML]{E9F7EF} 84.9 (+0.3)}&83.8&\multicolumn{2}{l}{\cellcolor[HTML]{E9F7EF} 84.2 (+0.4)}&83.4&\multicolumn{2}{l}{\cellcolor[HTML]{E9F7EF} 84.2 (+0.8)}\\
\textit{he}&56.0&\multicolumn{2}{l}{\cellcolor[HTML]{52BE80} 67.1 (+11.1)}&53.3&\multicolumn{2}{l}{\cellcolor[HTML]{52BE80} 63.3 (+10.1)}&57.9&\multicolumn{2}{l}{\cellcolor[HTML]{7DCEA0} 66.1 (+8.2)}&-&\multicolumn{2}{l}{-}\\
\textit{hi}&69.5&\multicolumn{2}{l}{\cellcolor[HTML]{7DCEA0} 75.7 (+6.2)}&70.1&\multicolumn{2}{l}{\cellcolor[HTML]{7DCEA0} 76.0 (+5.9)}&74.8&\multicolumn{2}{l}{\cellcolor[HTML]{A9DFBF} 77.1 (+2.2)}&76.0&\multicolumn{2}{l}{\cellcolor[HTML]{E9F7EF} 76.4 (+0.4)}\\
\textit{hu}&81.0&\multicolumn{2}{l}{\cellcolor[HTML]{FDEDEC} 80.7 (-0.3)}&76.0&\multicolumn{2}{l}{\cellcolor[HTML]{7DCEA0} 82.0 (+6.0)}&76.5&\multicolumn{2}{l}{\cellcolor[HTML]{A9DFBF} 80.0 (+3.5)}&-&\multicolumn{2}{l}{-}\\
\textit{id}&52.4&\multicolumn{2}{l}{\cellcolor[HTML]{A9DFBF} 53.9 (+1.5)}&77.6&\multicolumn{2}{l}{\cellcolor[HTML]{E9F7EF} 77.9 (+0.3)}&82.2&\multicolumn{2}{l}{\cellcolor[HTML]{E9F7EF} 82.3 (+0.1)}&-&\multicolumn{2}{l}{-}\\
\textit{it}&81.3&\multicolumn{2}{l}{\cellcolor[HTML]{FDEDEC} 80.9 (-0.4)}&86.2&\multicolumn{2}{l}{\cellcolor[HTML]{E9F7EF} 86.4 (+0.1)}&86.4&\multicolumn{2}{l}{\cellcolor[HTML]{FADBD8} 85.5 (-1.0)}&-&\multicolumn{2}{l}{-}\\
\textit{ja}&18.0&\multicolumn{2}{l}{\cellcolor[HTML]{52BE80} 43.5 (+25.5)}&28.3&\multicolumn{2}{l}{\cellcolor[HTML]{52BE80} 38.3 (+10.0)}&29.8&\multicolumn{2}{l}{\cellcolor[HTML]{7DCEA0} 38.0 (+8.3)}&-&\multicolumn{2}{l}{-}\\
\textit{jv}&62.3&\multicolumn{2}{l}{\cellcolor[HTML]{7DCEA0} 69.8 (+7.5)}&72.4&\multicolumn{2}{l}{\cellcolor[HTML]{A9DFBF} 75.7 (+3.2)}&72.9&\multicolumn{2}{l}{\cellcolor[HTML]{FDEDEC} 72.3 (-0.6)}&-&\multicolumn{2}{l}{-}\\
\textit{ka}&68.8&\multicolumn{2}{l}{\cellcolor[HTML]{7DCEA0} 74.7 (+5.9)}&60.6&\multicolumn{2}{l}{\cellcolor[HTML]{52BE80} 72.2 (+11.6)}&67.1&\multicolumn{2}{l}{\cellcolor[HTML]{7DCEA0} 72.5 (+5.4)}&-&\multicolumn{2}{l}{-}\\
\textit{kk}&53.5&\multicolumn{2}{l}{\cellcolor[HTML]{E9F7EF} 54.2 (+0.7)}&32.7&\multicolumn{2}{l}{\cellcolor[HTML]{52BE80} 53.1 (+20.4)}&26.1&\multicolumn{2}{l}{\cellcolor[HTML]{52BE80} 51.7 (+25.5)}&-&\multicolumn{2}{l}{-}\\
\textit{ko}&57.9&\multicolumn{2}{l}{\cellcolor[HTML]{A9DFBF} 61.9 (+4.0)}&33.7&\multicolumn{2}{l}{\cellcolor[HTML]{7DCEA0} 39.1 (+5.4)}&30.6&\multicolumn{2}{l}{\cellcolor[HTML]{52BE80} 44.7 (+14.1)}&-&\multicolumn{2}{l}{-}\\
\textit{ml}&63.5&\multicolumn{2}{l}{\cellcolor[HTML]{E9F7EF} 64.3 (+0.8)}&42.1&\multicolumn{2}{l}{\cellcolor[HTML]{52BE80} 65.1 (+23.0)}&42.5&\multicolumn{2}{l}{\cellcolor[HTML]{52BE80} 63.9 (+21.3)}&-&\multicolumn{2}{l}{-}\\
\textit{mr}&63.9&\multicolumn{2}{l}{\cellcolor[HTML]{A9DFBF} 67.1 (+3.2)}&49.6&\multicolumn{2}{l}{\cellcolor[HTML]{7DCEA0} 57.4 (+7.9)}&53.9&\multicolumn{2}{l}{\cellcolor[HTML]{A9DFBF} 55.6 (+1.8)}&-&\multicolumn{2}{l}{-}\\
\textit{ms}&64.1&\multicolumn{2}{l}{\cellcolor[HTML]{52BE80} 74.1 (+10.0)}&79.3&\multicolumn{2}{l}{\cellcolor[HTML]{E9F7EF} 79.6 (+0.3)}&80.5&\multicolumn{2}{l}{\cellcolor[HTML]{FADBD8} 79.4 (-1.1)}&-&\multicolumn{2}{l}{-}\\
\textit{my}&53.5&\multicolumn{2}{l}{\cellcolor[HTML]{7DCEA0} 62.0 (+8.5)}&35.0&\multicolumn{2}{l}{\cellcolor[HTML]{A9DFBF} 38.7 (+3.7)}&31.9&\multicolumn{2}{l}{\cellcolor[HTML]{A9DFBF} 33.0 (+1.1)}&-&\multicolumn{2}{l}{-}\\
\textit{nl}&84.1&\multicolumn{2}{l}{\cellcolor[HTML]{FADBD8} 83.1 (-1.0)}&84.2&\multicolumn{2}{l}{\cellcolor[HTML]{A9DFBF} 85.5 (+1.3)}&83.5&\multicolumn{2}{l}{\cellcolor[HTML]{E9F7EF} 84.1 (+0.5)}&-&\multicolumn{2}{l}{-}\\
\textit{pt}&82.0&\multicolumn{2}{l}{\cellcolor[HTML]{FADBD8} 80.1 (-1.9)}&83.0&\multicolumn{2}{l}{82.9 (+0.0)}&83.5&\multicolumn{2}{l}{\cellcolor[HTML]{FDEDEC} 82.7 (-0.8)}&-&\multicolumn{2}{l}{-}\\
\textit{ru}&71.1&\multicolumn{2}{l}{\cellcolor[HTML]{FADBD8} 68.2 (-2.9)}&55.3&\multicolumn{2}{l}{\cellcolor[HTML]{52BE80} 70.8 (+15.6)}&59.8&\multicolumn{2}{l}{\cellcolor[HTML]{52BE80} 70.1 (+10.3)}&65.6&\multicolumn{2}{l}{\cellcolor[HTML]{7DCEA0} 72.8 (+7.2)}\\
\textit{sw}&68.5&\multicolumn{2}{l}{\cellcolor[HTML]{A9DFBF} 70.7 (+2.2)}&65.9&\multicolumn{2}{l}{\cellcolor[HTML]{E9F7EF} 66.4 (+0.5)}&66.8&\multicolumn{2}{l}{\cellcolor[HTML]{7DCEA0} 73.9 (+7.1)}&-&\multicolumn{2}{l}{-}\\
\textit{ta}&58.8&\multicolumn{2}{l}{\cellcolor[HTML]{7DCEA0} 64.3 (+5.5)}&49.5&\multicolumn{2}{l}{\cellcolor[HTML]{52BE80} 61.7 (+12.1)}&52.6&\multicolumn{2}{l}{\cellcolor[HTML]{52BE80} 63.3 (+10.7)}&-&\multicolumn{2}{l}{-}\\
\textit{te}&55.6&\multicolumn{2}{l}{\cellcolor[HTML]{A9DFBF} 57.4 (+1.8)}&47.4&\multicolumn{2}{l}{\cellcolor[HTML]{52BE80} 57.5 (+10.1)}&51.3&\multicolumn{2}{l}{\cellcolor[HTML]{7DCEA0} 57.8 (+6.5)}&-&\multicolumn{2}{l}{-}\\
\textit{th}&0.7&\multicolumn{2}{l}{\cellcolor[HTML]{52BE80} 15.1 (+14.4)}&2.0&\multicolumn{2}{l}{\cellcolor[HTML]{A9DFBF} 3.8 (+1.8)}&2.0&\multicolumn{2}{l}{\cellcolor[HTML]{7DCEA0} 7.4 (+5.4)}&-&\multicolumn{2}{l}{-}\\
\textit{tl}&73.0&\multicolumn{2}{l}{\cellcolor[HTML]{7DCEA0} 80.0 (+7.0)}&80.6&\multicolumn{2}{l}{\cellcolor[HTML]{A9DFBF} 81.6 (+1.0)}&81.9&\multicolumn{2}{l}{\cellcolor[HTML]{A9DFBF} 83.2 (+1.3)}&-&\multicolumn{2}{l}{-}\\
\textit{tr}&80.3&\multicolumn{2}{l}{\cellcolor[HTML]{FDEDEC} 79.6 (-0.7)}&68.8&\multicolumn{2}{l}{\cellcolor[HTML]{A9DFBF} 69.7 (+1.0)}&71.4&\multicolumn{2}{l}{\cellcolor[HTML]{FADBD8} 68.8 (-2.6)}&-&\multicolumn{2}{l}{-}\\
\textit{ur}&63.6&\multicolumn{2}{l}{\cellcolor[HTML]{52BE80} 74.7 (+11.1)}&51.4&\multicolumn{2}{l}{\cellcolor[HTML]{52BE80} 65.4 (+14.0)}&56.9&\multicolumn{2}{l}{\cellcolor[HTML]{52BE80} 67.0 (+10.1)}&-&\multicolumn{2}{l}{-}\\
\textit{vi}&74.2&\multicolumn{2}{l}{\cellcolor[HTML]{A9DFBF} 76.0 (+1.8)}&81.4&\multicolumn{2}{l}{\cellcolor[HTML]{A9DFBF} 83.0 (+1.6)}&81.7&\multicolumn{2}{l}{\cellcolor[HTML]{E9F7EF} 82.0 (+0.4)}&82.4&\multicolumn{2}{l}{\cellcolor[HTML]{FADBD8} 79.6 (-2.8)}\\
\textit{yo}&37.1&\multicolumn{2}{l}{\cellcolor[HTML]{52BE80} 73.8 (+36.7)}&75.7&\multicolumn{2}{l}{\cellcolor[HTML]{7DCEA0} 82.3 (+6.6)}&75.5&\multicolumn{2}{l}{\cellcolor[HTML]{A9DFBF} 78.4 (+3.0)}&-&\multicolumn{2}{l}{-}\\
\textit{zh}&27.1&\multicolumn{2}{l}{\cellcolor[HTML]{52BE80} 45.9 (+18.8)}&31.1&\multicolumn{2}{l}{\cellcolor[HTML]{7DCEA0} 39.7 (+8.7)}&31.6&\multicolumn{2}{l}{\cellcolor[HTML]{7DCEA0} 39.8 (+8.2)}&36.3&\multicolumn{2}{l}{\cellcolor[HTML]{7DCEA0} 43.2 (+6.9)}\\
\midrule
\textit{AVG}&64.3&\multicolumn{2}{l}{\cellcolor[HTML]{A9DFBF} 68.9 (+4.6)}&61.2&\multicolumn{2}{l}{\cellcolor[HTML]{7DCEA0} 68.5 (+7.4)}&62.9&\multicolumn{2}{l}{\cellcolor[HTML]{7DCEA0} 68.6 (+5.7)}&71.0&\multicolumn{2}{l}{\cellcolor[HTML]{A9DFBF} 73.3 (+2.3)}\\
\bottomrule
\end{tabular}
\vspace{-5pt}
\caption{Cross-lingual NER F$_1$ on WikiANN for mT5 and XLM-RoBERTa$_{\text{large}}$. Due to the computing limit, we run the largest mT5$_{\text{XXL}}$ model on 8 languages which were chosen following~\citet{wu-2021-everything}. The performance is averaged over 3 runs for XLM-R$_{\text{large}}$ and mT5$_{\text{large}}$ models, and  1 run for mT5$_{\text{XL}}$ and mT5$_{\text{XXL}}$ models. Models are fine-tuned on a combination of English and EasyProject data with Google Translation.}
\label{table:mt5_detail}
\end{table*}

\end{document}